%% file: SiamGM_KBS.tex
\documentclass[a4paper,fleqn]{cas-dc}

\usepackage[numbers,sort&compress]{natbib}
\usepackage{algorithm}
\usepackage{algorithmic}
\usepackage[caption=false]{subfig}
\usepackage{tabularx}

\hypersetup{
  colorlinks=true,
  linkcolor=blue,
  citecolor=blue,
  urlcolor=blue
}

\graphicspath{{figs/}}

\newcolumntype{Y}{>{\centering\arraybackslash}X}

\begin{document}

\let\WriteBookmarks\relax
\def\floatpagepagefraction{1}
\def\textpagefraction{.001}

\shorttitle{Geometric-Topological Perception and Motion Prior for Real-Time SVOT}
\shortauthors{Z. Wen et al.}

\title[mode=title]{Geometric-Topological Perception and Motion Prior for Real-Time Satellite Video Object Tracking}

\author[1,2,3]{Zixiao Wen}[orcid=0009-0008-3729-0538]
\ead{wenzixiao22@mails.ucas.ac.cn}

\author[1,2,3]{Guangyao Zhou}
\ead{zhougy@aircas.ac.cn}

\author[1,2,3]{Jiawei Li}
\ead{lijiawei231@mails.ucas.ac.cn}

\author[1,2,3]{Xiantai Xiang}
\ead{xiangxiantai@gmail.com}


\author[1,2]{Zhen Yang}
\ead{yangzhen003999@aircas.ac.cn}

\author[1,2,3]{Yuxin Hu}
\ead{huyx@aircas.ac.cn}

\author[1,2,3]{Yuhan Liu}
\cormark[1]
\ead{liuyuhan@aircas.ac.cn}

\affiliation[1]{
  organization={Aerospace Information Research Institute, Chinese Academy of Sciences},
  city={Beijing},
  postcode={100094},
  country={China}
}

\affiliation[2]{
  organization={Key Laboratory of Technology in Geo-Spatial Information Processing and Application System, Chinese Academy of Sciences},
  city={Beijing},
  postcode={100190},
  country={China}
}

\affiliation[3]{
  organization={School of Electronic, Electrical and Communication Engineering, University of Chinese Academy of Sciences},
  city={Beijing},
  postcode={101408},
  country={China}
}


\cortext[1]{Corresponding author.}

\begin{abstract}
Satellite video object tracking (SVOT) remains fundamentally challenging due to texture scarcity, arbitrary rotation, aspect ratio changes, and severe occlusions.
While recent state-of-the-art trackers excel in general scenarios, their reliance on rich appearance details or rigid spatial matching mechanisms leads to significant performance degradation in the satellite domain.
To bridge this gap, we propose SiamGM, a real-time spatial-temporal unified tracking framework tailored for satellite videos.
Instead of conventionally stacking modules, we synergize geometric-topological perception with temporal-kinematic prior, addressing the core limitations of SVOT.
Spatially, we closely couple a Topological Attention Module (TAM) with a Geometry-Constrained Label Assignment (GCLA) method during the training phase, where a Local Graph Propagation (LGP) mechanism enforces neighborhood-consistent template--search correspondences.
This joint paradigm establishes fine-grained structural correspondences and explicitly suppresses surrounding background noise.
Temporally, rather than passively relying on visual cues, we propose an Online Motion Model Refinement (OMMR) strategy during the tracking phase.
Gated by an online self-calibrated confidence indicator, OMMR adaptively fuse long-term stable and short-term instantaneous motion estimation for robust tracking.
Evaluations on two challenging SatSOT and SV248S benchmarks confirm that SiamGM outperforms not only recent Siamese-based satellite trackers but also state-of-the-art Transformer trackers in both precision and success metrics.
Notably, this highly unified architecture introduces only minor computational overhead, enabling real-time tracking at 130 frames per second (FPS).
\end{abstract}



\maketitle

\input{introduction}
\input{related_work}
\input{proposed_method}
\input{experiment}
\input{discussion}
\input{conclusion}

\bibliographystyle{cas-model2-names}
\bibliography{reference}

\end{document}

%% file: introduction.tex
\section{Introduction}
\label{sec:introduction}

Satellite video object tracking (SVOT) has emerged as a critical research area in computer vision, driven by the increasing availability of spaceborne remote sensing imagery and the growing demand for real-time applications such as intelligent surveillance, environmental monitoring, and disaster response~\cite{chen2024satellite}.
Among them, single object tracking (SOT), as an important branch, aims to continuously track a specific target throughout satellite videos (SVs). 
Unlike multiple object tracking (MOT), SOT requires exceptionally higher robustness and efficiency.
Any momentary tracking failure can irreversibly collapse the subsequent trajectory, which is intolerable for critical real-world tasks such as suspicious target tracking and dynamic situation awareness.

\begin{figure}[!t]
    \centering
    \subfloat[]{
        \includegraphics[width=0.47\linewidth]{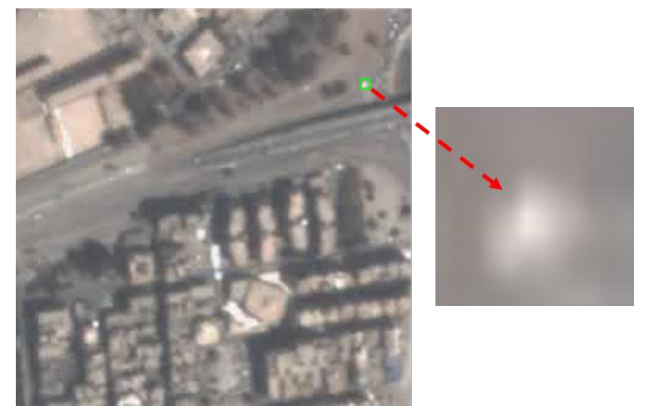}
        \label{fig:challenge1}
    }\hfill
    \subfloat[]{
        \includegraphics[width=0.47\linewidth]{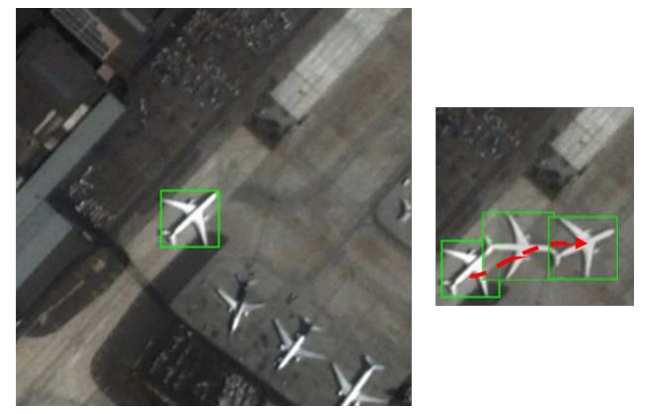}
        \label{fig:challenge2}
    }\\[-0.5em]
    \subfloat[]{
        \includegraphics[width=0.47\linewidth]{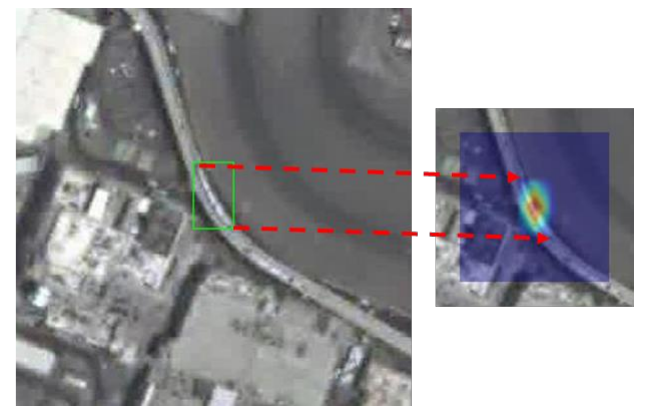}
        \label{fig:challenge3}
    }\hfill
    \subfloat[]{
        \includegraphics[width=0.47\linewidth]{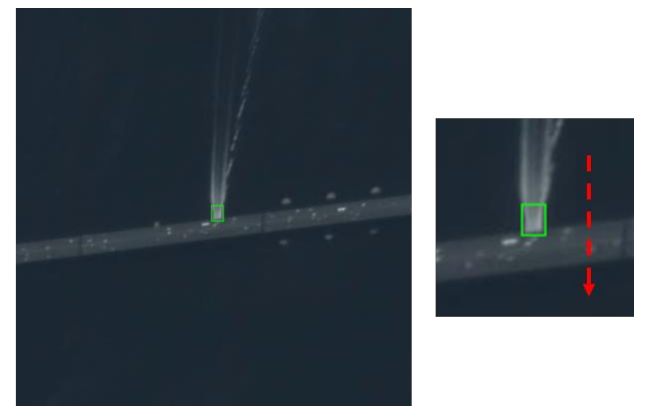}
        \label{fig:challenge4}
    }
    \caption{Visual illustrations of the inherent challenges in satellite videos. (a) Tiny objects lack distinct texture features. (b) Targets often undergo arbitrary rotations, causing misalignment between search and template features. (c) Non-rigid targets undergo non-linear structural deformations, leading to large aspect ratio change and severe background noise within horizontal bounding boxes. (d) Severe partial or full occlusions easily lead to tracking drift.}
    \label{fig:challenges}
    \vspace{-5pt}
\end{figure}

Despite the remarkable success of modern trackers in natural scenes, their direct transfer to SVOT reveals a fundamental conceptual misalignment.
Compared to standard sequences, SVs present far more extreme and unique challenges, as visually demonstrated by the representative scenarios and targets in Fig.~\ref{fig:challenges}.
At the micro-level, satellite targets typically occupy only a few dozen pixels~\cite{wen2025fanet,wen2026d3rdetr}, leading to texture scarcity and blurred boundaries and completely stripping away reliable appearance cues, as shown in Fig.~\ref{fig:challenges}\subref{fig:challenge1}.
Compounded by the inherent top-down perspective, targets frequently undergo arbitrary rotations during movement, as illustrated in Fig.~\ref{fig:challenges}\subref{fig:challenge2}, causing traditional spatial-aligned convolutions to severely struggle in achieving high-correlation feature matching.
At the macro-level, besides rotating scenes, targets like multi-carriage trains act as non-rigid objects when navigating curved tracks, undergoing severe non-linear structural deformations.
Consequently, traditional horizontal axis-aligned bounding boxes fail to tightly enclose these objects, which is evident in Fig.~\ref{fig:challenges}\subref{fig:challenge3}.
When combined with standard centrally-symmetric label assignment, massive background pixels are inevitably allocated high positive weights (as shown in the heatmap on the right), injecting severe false alarms that disrupt boundary regression.
These challenges can be fundamentally attributed to the mismatch between rigid spatial appearance modeling and the dynamic, structure-varying nature of targets in satellite videos.

Even if a tracker achieves flawless spatial feature matching, SVOT inevitably confronts an ultimate physical barrier in the temporal dimension: continuous partial or full occlusions (e.g., targets entering thick clouds or passing under massive bridges, as in Fig.~\ref{fig:challenges}\subref{fig:challenge4}).
Under such severe circumstances, all visual features are rendered entirely incapacitated, ultimately leading to severe and irreversible tracking drift.
Fortunately, the fixed top-down perspective of SVs provides a unique kinematic prior: the physical tracking trajectories are remarkably smooth and predictable, which indicates that dynamically bridging visual gaps with historical temporal motion prior is the final defense for robust tracking.

Most state-of-the-art models, especially heavy transformer-based trackers typically rely on rigid spatial tokenization or rich appearance and fine-grained texture details to track targets.
Conversely, while pure Convolutional Neural Networks (CNNs) inherently preserve high-resolution local details essential for satellite targets, their traditional cross-correlation matching paradigm acts as a rigid sliding window, which is fundamentally misaligned with the severe spatial transformations in SVs.
Our key insight is that robust satellite video object tracking requires jointly modeling shape-aware spatial correspondence and motion-consistent temporal evolution, rather than relying on either aspect alone.
To systematically overcome these limitations, we propose SiamGM, a real-time spatial-temporal unified tracking framework tailored specifically for SVs.
SiamGM systematically tackles the aforementioned challenges from spatial geometric-topological and temporal motion perspectives.
Spatially, to overcome the vulnerability of degraded appearance features, we transition from isolated point-wise matching to structure-aware correspondence modeling.
We formulate a Topological Attention Module (TAM) that represents the template and search features as grid graphs and propagates correspondence evidence via a Local Graph Propagation (LGP) mechanism, inherently resists severe spatial transformations.
Concurrently, to eliminate spatial ambiguities caused by unconstrained target shapes, we embed macroscopic physical geometry into the network optimization via a Geometry-Constrained Label Assignment (GCLA) formulation.
This mechanism dynamically restricts positive sampling strictly along the moving target's principal axis, fundamentally resolving boundary degradation.
Temporally, to tackle occlusions where all spatial visual features fail, we propose the Online Motion Model Refinement (OMMR) strategy, an online refinement scheme with motion prior.
OMMR decouples temporal perception into a stable long-term window and an instantaneous short-term window, and switches between them according to a self-calibrated confidence, rectifying erroneous visual predictions and safely maintain trajectory continuity during severe or complete occlusions.
The main contributions of this work are summarized as follows:
\begin{enumerate}
    \item We tackle the fundamental limitations of existing SVOT methods by proposing SiamGM, a spatial-temporal unified tracking framework with geometric-topological perception and motion prior.
    \item Spatially, we introduce the TAM equipped with the LGP mechanism, which enforces topological correspondences under rotation and deformation with blurred features, and is seamlessly coupled with the GCLA method.
    \item Temporally, we propose the OMMR strategy for the tracking phase, a confidence-gated dual-frequency motion refinement scheme that safely exploits historical trajectory information for robust tracking under occlusion and long-term disappearance.
    \item Extensive experiments on two large-scale challenging SatSOT and SV248S benchmarks demonstrate that our method achieves state-of-the-art performance while maintaining real-time efficiency.
\end{enumerate}

The remainder of this paper is organized as follows.
Section~\ref{sec:related_work} briefly reviews the related work. 
Section~\ref{sec:proposed_method} elaborates on the architectural details of the proposed SiamGM framework.
Section~\ref{sec:experiment} presents the comprehensive experimental results and ablation studies, followed by an in-depth discussion and visual analysis in Section~\ref{sec:discussion}.
Finally, Section~\ref{sec:conclusion} concludes this paper and outlines future research directions.

%% file: related_work.tex
\section{Related Work}
\label{sec:related_work}

\subsection{Deep Learning-Based Single Object Tracking}

In recent years, Siamese neural networks (SNNs) have achieved dominant performance in the field of SOT due to their balanced accuracy and efficiency.
The pioneering work, SiamFC~\cite{bertinetto2016fully}, formulated tracking as a similarity learning problem via cross-correlation between feature maps.
By sharing weights and utilizing multi-scale representations, it realizes highly efficient online tracking.
Following this, SiamRPN~\cite{li2018high} and SiamRPN++~\cite{li2019siamrpn++} introduced the Region Proposal Network (RPN)~\cite{ren2015faster} to simultaneously perform classification and bounding box regression. 
Subsequent trackers, such as DaSiamRPN~\cite{zhu2018distractor}, SA-Siam~\cite{he2018twofold}, and C-RPN~\cite{fan2019siamese}, extended this paradigm, often employing architectures like AlexNet to strike a favorable balance between tracking speed and precision.
To eliminate the parameter tuning associated with reference boxes, SiamCAR~\cite{guo2020siamcar} pioneered a fully convolutional anchor-free Siamese framework.
It decomposes tracking into pixel-wise classification and regression, introducing a centerness branch to effectively suppress low-quality predictions.
Building upon this elegant design, numerous subsequent studies~\cite{shan2020siamfpn, zhang2021siamcda, bao2022siamese, zheng2022leveraging, bao2023siamthn} further solidified the dominance of Siamese architectures in SOT.

Recently, Transformer-based trackers have emerged, combining the global modeling capabilities of attention mechanisms with the efficient feature extraction of Convolutional Neural Networks (CNNs) to further push the performance boundaries (e.g., TransT~\cite{chen2021transformer}, TrDiMP~\cite{wang2021transformer}, and STARK~\cite{yan2021learning}).
Furthermore, visual fully-transformer trackers gradually evolved into two major architectural paradigms.
The two-stream two-stage paradigm utilizes separate Transformers for independent feature extraction and subsequent information interaction, including DualTFR~\cite{xie2021learning}, SwinTrack~\cite{lin2022swintrack} and others.
Conversely, the one-stream one-stage paradigm, such as OSTrack~\cite{ye2022joint}, MixFormer~\cite{cui2022mixformer}, SeqTrack~\cite{chen2023seqtrack}, and ARTrack~\cite{wei2023autoregressive} etc., elegantly unifies feature extraction and relation modeling within a single unified Transformer backbone.

Despite their remarkable success in general scenarios, these trackers implicitly assume that targets are relatively compact and axis-aligned, lacking targeted geometric awareness.
This assumption inherently limits their adaptability and causes severe performance degradation when confronted with rotating objects and the extreme aspect ratios typically found in remote sensing imagery.

\subsection{Single Object Tracking in Satellite Videos}

SVOT presents unique challenges distinct from natural scenes, including small target scales, similar distractors, and complex background clutter.
Early approaches relied on Correlation Filters (CF) due to their computational efficiency, such as CFME~\cite{xuan2019object}, CFKF~\cite{guo2019object}.
However, CF-based trackers often struggle with severe deformation and semantic variations.
Recently, deep learning-based methods have been tailored for SVs.
To address feature representation issues, techniques incorporating multi-level correlations have gained attention.
Lai et al.~\cite{lai2023target} proposed a bidirectional propagation and fusion (Bi-PF) module to enhance feature representation.
Wang et al.~\cite{wang2024satellite} proposed SVLPNet to solve the textureless and low-resolution issues by introducing location prompts.
Zhang et al.~\cite{zhang2024adaptive} proposed an Adaptive Multi-Scale Transformer (MT) to capture multi-scale features and adaptively fuse them for robust tracking.
Wang et al.~\cite{wang2024contrastive} introduced CAMTracker, which incorporates a contrastive-augmented matching module to enhance feature discriminability.
To better classify small targets, Xue et al.~\cite{xue2023smalltrack} introduced SmallTrack to introduce a graph enhanced module, and Yang et al.~\cite{yang2024ehtracker} proposed EHTracker to enhance the discriminability of small targets by refining the classification feature map through multihead graph attention.
To utilize temporal information, some methods have explored temporal modeling.
Yang et al.~\cite{yang2023siammdm} proposed SiamMDM with an dynamic template update strategy to mitigate the impact of background clutter.
Lv et al.~\cite{lv2025siams2f} introduced SiamS\textsuperscript{2}F, employing spectral-spatial-frame correlation to enhance feature discriminability in hyperspectral satellite videos.
Zhou et al.~\cite{zhou2025siamtitp} proposed SiamTITP, which integrates a temporal memory module and trajectory prediction to handle occlusion and target disappearance.

While these methods have improved robustness against small scales and occlusions, they largely ignore the fine-grained topological prior and geometric constraints during the label assignment phase.
Consequently, boundary regression degrades significantly for severely elongated satellite targets.

\subsection{Motion Modeling in Satellite Video Object Tracking}

Motion modeling is crucial for SVOT, especially for handling occlusions and predicting target trajectories.
Traditional methods often rely on Kalman filters or particle filters to estimate motion, but they can be limited in complex scenarios, such as CFKF~\cite{guo2019object} and Siam-TMC~\cite{nie2022object}.
Recent approaches have incorporated deep learning-based motion models.
For example, Li et al.~\cite{li2022object} introduced an interacting mutiple model (IMM) through rotation state transition.
Ruan et al.~\cite{ruan2022deep} proposed SRN-TFM which fitting trajectory with a polynomial function to predict the future position of the target.
Yang et al. \cite{yang2023siammdm} proposed a motion model that refines the search area based on the predicted motion, improving tracking performance during occlusions.
Zhou et al.~\cite{zhou2025incorporating} integrated a trajectory prediction module to capture long-term dependencies using linear fitting, enhancing tracking performance during occlusions.
Furthermore, several learning-based approaches, such as MBLT~\cite{zhang2021mblt} and AD-OHNet~\cite{cui2021remote}, employ CNNs or reinforcement learning strategies to analyze multi-frame trajectory patterns and forecast future positions.
While these methods enhance temporal continuity, they often introduce substantial computational overhead.
Furthermore, many existing methods lack a dynamic reliability assessment mechanism, indiscriminately fusing temporal memory even when current visual observations are severely corrupted, inevitably leading to tracking drift.

To fundamentally bridge these spatial and temporal gaps, this paper proposes SiamGM, a novel tracking framework with geometric-topological perception and motion prior.
Unlike previous works, we explicitly establish structural correspondences via TAM and regulate specific geometric proportions using the GCLA method.
Moreover, avoiding cumbersome recurrent networks, we introduce the OMMR strategy, a lightweight confidence-gated motion refinement scheme.
This elegant combination systematically ensures real-time robustness under extreme structural distortions and severe occlusions.

%% file: proposed_method.tex
\section{Proposed Method}
\label{sec:proposed_method}

\begin{figure*}[!t]
    \centering
    \includegraphics[width=0.95\linewidth]{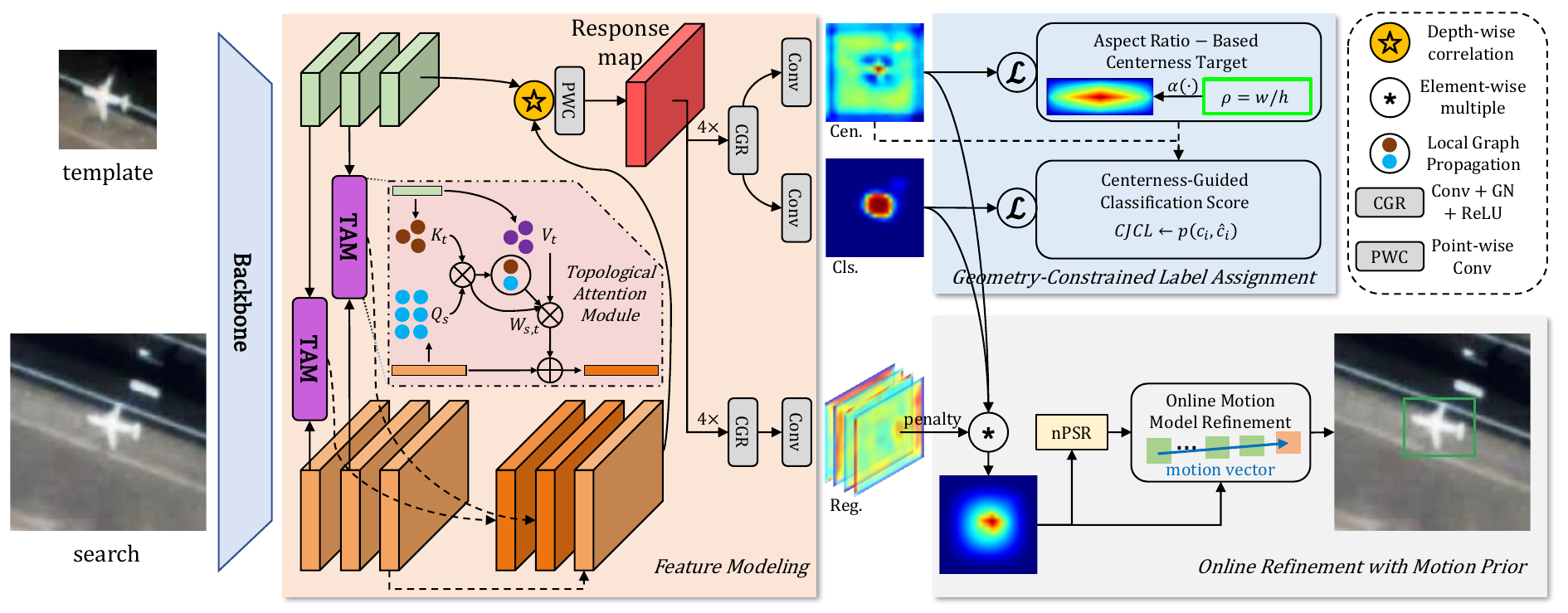}
    \caption{Overall architecture of the proposed SiamGM framework. The Topological Attention Module (TAM) is integrated to establish fine-grained, topology-consistent correspondences via the Local Graph Propagation (LGP) mechanism over the feature grid graphs. Geometry-Constrained Label Assignment (GCLA) is employed during the training phase to explicitly supervise spatial ambiguities. The Online Motion Model Refinement (OMMR) strategy is deployed as a post-processing stage to dynamically rectify trajectory bounds based on historical motion information. CGR denotes the combination of convolution, group normalization, and ReLU activation, while PWC denotes point-wise convolution.}
    \label{fig:overview}
    \vspace{-5pt}
\end{figure*}

The overall architecture of the proposed SiamGM framework is illustrated in Fig.~\ref{fig:overview}, systematically unifying geometric-topological perception formulations and online refinement with motion prior within an anchor-free pipeline.
Specifically, after hierarchical feature extraction, the TAM is integrated, in which the LGP mechanism refines the initial template--search correspondences by propagating matching evidence over local grid-graph neighborhoods.
Within the prediction head, the GCLA method is employed during the training phase to strictly supervise spatial ambiguities.
Finally, during the online tracking phase, the OMMR strategy is deployed as a post-processing stage to dynamically rectify trajectory bounds based on historical motion information.
Despite these structural enhancements, all integrated components introduce practically negligible computational overhead, ensuring that SiamGM operates robustly as a real-time framework.
The detailed formulations of these components are elaborated in the subsequent subsections.

\subsection{Spatial Geometric-Topological Perception}

In standard SNN-based trackers, template and search features are typically fused using depth-wise cross-correlation to generate response maps.
This rigid, sliding-window matching paradigm implicitly assumes that targets maintain stable appearances and consistent spatial alignments across frames.
However, targets in satellite videos inherently suffer from severe microscopic feature degradation (texture scarcity) and complex macroscopic spatial transformations (arbitrary rotations and aspect ratio changes), causing conventional spatial-aligned convolutions to fail catastrophically.
To address this issue, we systematically decouple the spatial challenge into two complementary perspectives: micro-level topological matching and macro-level geometric constraint.


\subsubsection{Topological Attention Module}

Targets in satellite videos usually occupy only a few pixels and provide limited texture cues.
Consequently, point-wise template--search similarities can become ambiguous and may produce isolated correspondences that are inconsistent with the surrounding target structure.
Therefore, we introduce the TAM to establish fine-grained structural correspondences by propagating correspondence evidence over the local neighborhoods in both graphs.

Given the search feature $F_s \in \mathbb{R}^{C \times H_s \times W_s}$ and the template feature $F_t \in \mathbb{R}^{C \times H_t \times W_t}$, we first generate query, key, and value representations using point-wise convolutions.
After flattening the spatial dimensions, we obtain $Q_s \in \mathbb{R}^{N_s \times d}$, $K_t \in \mathbb{R}^{d \times N_t}$, and $V_t \in \mathbb{R}^{C \times N_t}$, where $d=C/r$ ($r=4$ by default), $N_s=H_sW_s$, and $N_t=H_tW_t$.
The initial template--search affinity and its row-wise normalized attention are computed as:
\begin{equation}
S_{s,t}=Q_sK_t, \qquad
W_{s,t}^{(0)}=\operatorname{Softmax}_{t}(S_{s,t}),
\label{eq:base_spatial_attention}
\end{equation}
where the softmax is applied over the $N_t$ template locations for each search location.
Thus, $W_{s,t}^{(0)}(i,j)$ measures the appearance-based correspondence between the $i$-th search node and the $j$-th template node.

We then construct two fixed grid graphs $\mathcal{G}_s=(\mathcal{V}_s,\mathcal{E}_s)$ and $\mathcal{G}_t=(\mathcal{V}_t,\mathcal{E}_t)$ from the spatial layouts of $F_s$ and $F_t$, respectively.
Each node is connected to its valid $N$-neighborhood and itself. Let $A_x$ be the $N$-neighbor adjacency matrix for $x\in\{s,t\}$. Its row-normalized form with self-loops is
\begin{equation}
\bar{A}_x=D_x^{-1}(A_x+I), \qquad
D_x(i,i)=\sum_j(A_x+I)_{ij}.
\label{eq:normalized_adjacency}
\end{equation}
The initial correspondence matrix is then propagated over the search and template graphs via LGP as
\begin{equation}
R_{s,t}=\bar{A}_s W_{s,t}^{(0)}\bar{A}_t^{T}.
\label{eq:topology_propagation}
\end{equation}
This operation averages correspondence evidence over the valid local neighborhoods on both sides.
The final topological attention is obtained by injecting the propagated evidence into the original affinity logits:
\begin{equation}
W_{s,t}=\operatorname{Softmax}_{t}\!\left(S_{s,t}+\beta R_{s,t}\right),
\label{eq:spatial_attention}
\end{equation}
where $\beta$ is a learnable scalar initialized to zero.
The template information is subsequently aggregated as $\hat{F}_s=V_tW_{s,t}^{T}$ and reshaped to $C\times H_s\times W_s$.
A residual connection produces the enhanced search feature:
\begin{equation}
\label{eq:ifga_residual}
F_{s}^{out} = F_s + \gamma \hat{F}_s
\end{equation}
where $\gamma$ is another learnable scalar initialized to zero.
Here we set $N=4$ by default in LGP, and considering the computational efficiency and semantic levels of features, we only apply the TAM to the P2 and P3 feature outputs, maintaining overall real-time performance.



\subsubsection{Geometry-Constrained Label Assignment}

For each point $t_{(i,j)}=(l,r,t,b)$ on the feature map, the traditional centerness assignment for positive sample is calculated as:
\begin{equation}
\label{eq:centerness_assignment}
c^{o}_{(i,j)} = \sqrt{\frac{\min(l,r)}{\max(l,r)} \cdot \frac{\min(t,b)}{\max(t,b)}}
\end{equation}
where $l, r, t, b$ denote the distances from the corresponding point to the four sides of ground-truth box.
This formulation constrains the feature distribution based on the distances to the box boundaries.
However, in satellite videos, horizontal bounding boxes often fail to capture the true feature distribution for targets with large aspect ratios, where the principal axis contains more target information and the other axis may include background noise.
To address this, we introduce an aspect ratio-based modulation factor $\alpha$ and corresponding centerness label calculation as follows:
\begin{equation}
\label{eq:aspect_ratio_constraint}
\alpha(\rho) = \min(1,\rho^\gamma)
\end{equation}
\begin{equation}
\label{eq:aspect_ratio_constrained_centerness}
c_{(i,j)} = \sqrt{\left (\frac{\min(l,r)}{\max(l,r)}\right )^{\alpha(\frac{1}{\rho})} \cdot \left (\frac{\min(t,b)}{\max(t,b)}\right )^{\alpha(\rho)}}
\end{equation}
where $\gamma$ is a hyperparameter that controls the constraint strength of the aspect ratio $\rho$.
The following Fig.~\ref{fig:centerness_visualization} illustrates the centerness assignment strategies for two different methods when dealing with targets of varying aspect ratios.
Specifically, Fig.~\ref{fig:centerness_visualization}\subref{fig:centerness_1d_multi_gamma} shows how varying the hyperparameter $\gamma$ affects the centerness distribution along the principal axis: a larger $\gamma$ results in a slower decay, concentrating the centerness more along the main axis.
Fig.~\ref{fig:centerness_visualization}\subref{fig:centerness_2d_comparison} compares the traditional centerness assignment and our method with $\gamma=0.5$ for targets with large aspect ratios.
It can be observed that for elongated targets, the traditional strategy tends to assign positive samples to a larger portion of the background, leading to less accurate sample selection.
Our method dynamically adjusts the ground truth labels for the centerness branch to more accurately reflect the feature-aware region along the principal axis of the target.

\begin{figure}[!t]
    \centering
    \subfloat[]{%
        \includegraphics[width=0.5\linewidth]{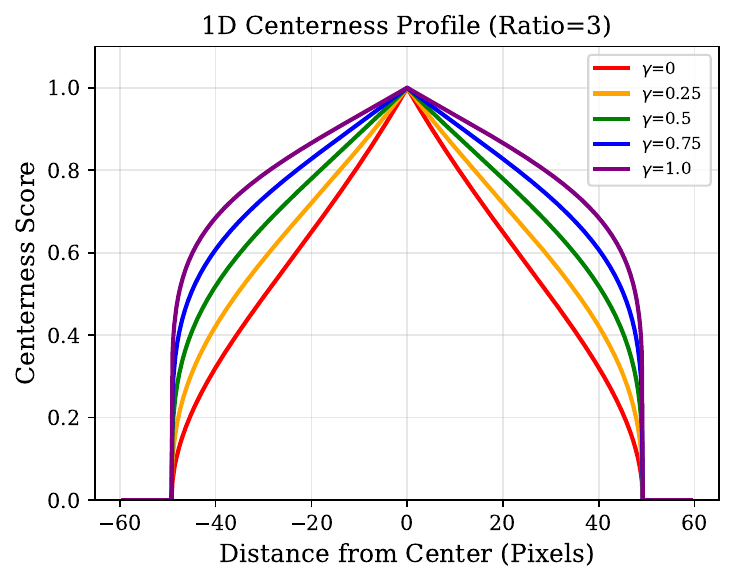}
        \label{fig:centerness_1d_multi_gamma}
    }\hfill
    \subfloat[]{%
        \includegraphics[width=0.45\linewidth]{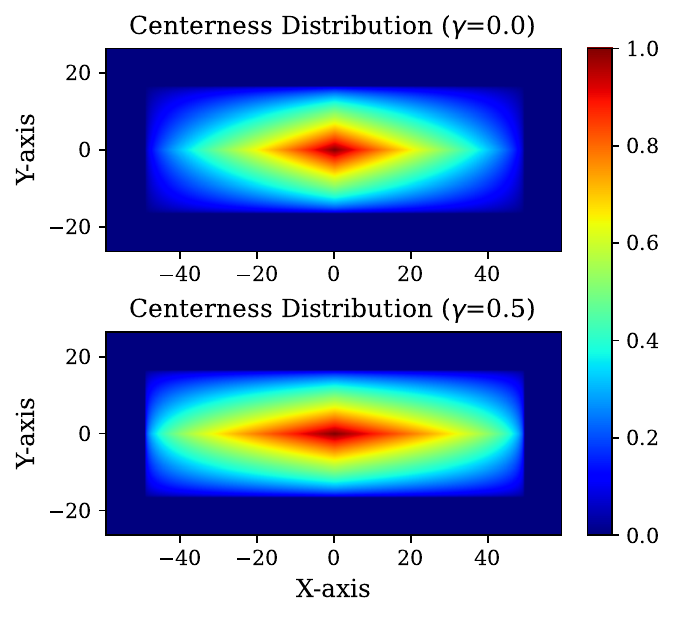}
        \label{fig:centerness_2d_comparison}
    }
    \caption{Visualization of (a) 1D multi-gamma centerness comparison and (b) 2D centerness distribution in which the aspect ratio is set to 3.}
    \label{fig:centerness_visualization}
    \vspace{-5pt}
\end{figure}


Moreover, although the classification and centerness branches are derived from the same feature representation during training, their loss functions are optimized independently.
However, during tracking, the confidence score is calculated as the product of the classification score and the centerness score, resulting in an inconsistency between the training and inference stages.
Consequently, the classification branch cannot fully utilize the spatial information provided by the centerness branch.
To address this limitation, we introduce a centerness-guided classification score (CGCS) to better reflect the reliability of each anchor during training.
Different from classification score in QFL~\cite{li2020generalized} and VFL~\cite{zhang2021varifocalnet}, our CGCS is specifically designed for the tracking task combined with centerness branch, defines the quality of anchor points based on both the predicted and target centerness values:
\begin{equation}
\label{eq:confidence_score}
p_{(i,j)} =
\begin{cases}
\frac{\min(c_{(i,j)},\hat{c}_{(i,j)})}{\max(c_{(i,j)},\hat{c}_{(i,j)})}, & \text{when } y_{(i,j)} = 1 \\
0, & \text{when } y_{(i,j)} = 0
\end{cases}
\end{equation}
where $\hat{c}_{(i,j)}$ is the predicted centerness score, $c_{(i,j)}$ is the aspect ratio-constrained centerness target, and $y_{(i,j)} \in \{0, 1\}$ denotes the binary classification label.
Note that $c_{(i,j)}$ is defined only within the positive sample region; for negative samples, we set $p_{(i,j)} = 0$ to suppress irrelevant background regions.
By introducing this score, we soften the traditional classification labels, allowing the predicted sample quality to be dynamically reflected in the classification process.

\subsubsection{Centerness-Joint Classification Loss}

Based on the proposed classification score below, we replace the BCELoss with a novel centerness-joint classification loss (CJCL), which guides the optimization of the classification branch, defined as follows:
\begin{equation}
\label{eq:cjcl}
\mathcal{L}_{CJCL} = -\frac{1}{N} \sum_{i=1}^{N} \left[ p_i\log(\hat{y}_i) + (1-p_i)\log(1-\hat{y}_i) \right]
\end{equation}
where $p_i$ is the CGCS for the $i$-th sample, and $\hat{y}_i$ is the corresponding prediction.
By weighting the classification loss with the CGCS, our method effectively suppresses noise in ambiguous regions, enhances the discriminative power of the classification scores, and ultimately improves tracking accuracy.

\subsection{Temporal Online Refinement with Motion Prior}

In satellite videos, typical scenarios such as vehicles passing under bridges, ships entering harbors frequently cause trackers to temporarily lose the target, resulting in short-term tracking failure, drift, or even complete loss due to occlusion, which is a common challenge in SVOT.
However, unlike general video sequences, satellite videos are captured by spaceborne sensors with relatively fixed fields of view and consistent spatial resolution, resulting in minimal viewpoint variation and negligible target deformation.
This unique physical mechanism dictates powerful temporal motion prior, that target movements in satellite videos typically follow strictly smooth, continuous, and predictable macroscopic trajectories, even when instantaneous visual cues are interrupted.
Leveraging this strong temporal prior, we transition from relying solely on isolated single-frame observations to introducing the OMMR strategy.
By dynamically evaluating the reliability of instantaneous visual responses, this strategy seamlessly bridges visual occlusion gaps using historical trajectory guidelines, recovering tracking continuity and robustness in challenging scenarios.

\subsubsection{Self-Calibrated Response Reliability}

\begin{figure}[!t]
    \centering
    \subfloat[]{%
        \includegraphics[width=0.47\linewidth]{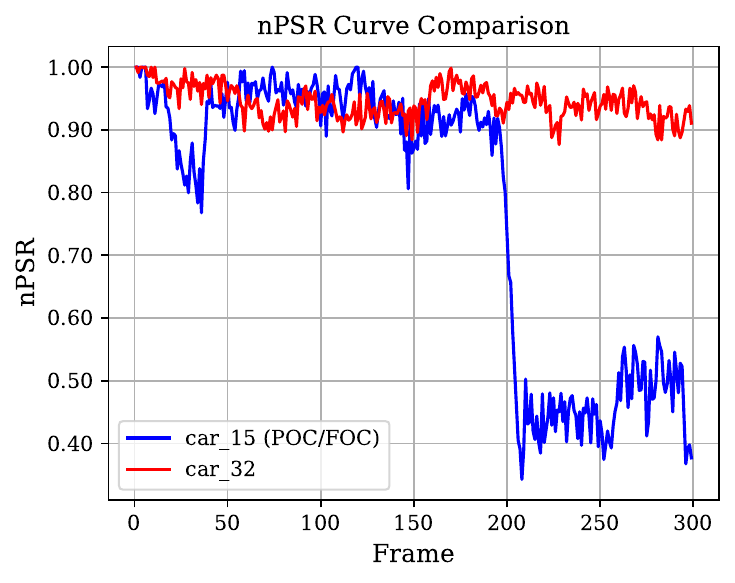}
        \label{fig:npsr_car}
    }\hfill
    \subfloat[]{%
        \includegraphics[width=0.47\linewidth]{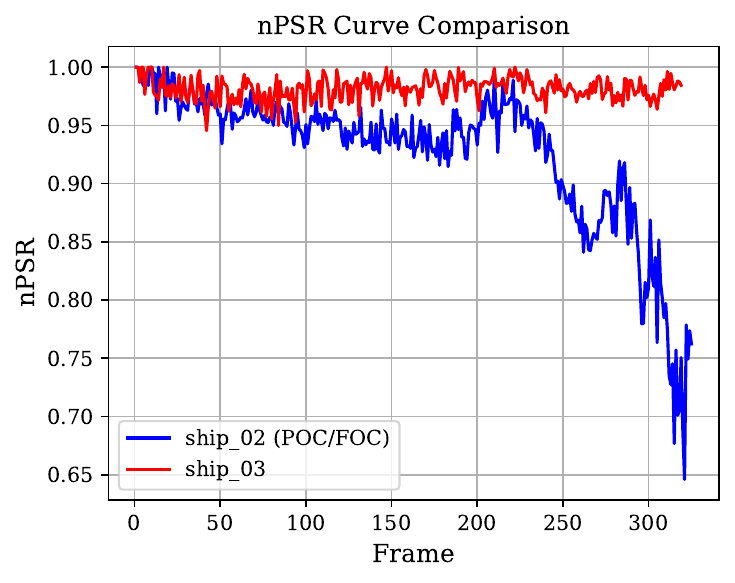}
        \label{fig:npsr_ship}
    }
    \caption{Visualization of nPSR curves for (a) two car targets and (b) two ship targets in SatSOT Dataset. The blue curve represents the nPSR values in normal scenarios, while the red curve corresponds to occlusion scenarios.}
    \label{fig:npsr_visualization}
    \vspace{-5pt}
\end{figure}

In each frame, the conventional tracker determines the target position by locating the peak value in the response score map.
However, in challenging scenarios, the true target location may be obscured by multiple competing peaks of similar magnitude, making accurate target localization difficult.
To determine when historical motion priors should take over, OMMR requires a per-frame reliability measure of the response score map.
The Peak-to-Sidelobe Ratio (PSR)~\cite{bolme2010visual} is a widely adopted measure for this purpose.
However, its magnitude varies significantly with different targets and backgrounds, we therefore calibrate the PSR online by its running maximum:
\begin{align}
\text{PSR} &= \frac{g_{\text{max}} - \mu_{\text{sl}}}{\sigma_{\text{sl}}} \label{eq:psr} \\
\text{nPSR} &= \frac{\text{PSR}}{\text{PSR}_{\text{max}}} \label{eq:npsr}
\end{align}
where $g_{\text{max}}$ denotes the peak value of the response score map, $\mu_{\text{sl}}$ and $\sigma_{\text{sl}}$ are the mean and standard deviation of the sidelobe region, respectively.
$\text{PSR}_{\text{max}}$ is a dynamically maintained variable updated throughout the online tracking process, used for normalization.
This online self-calibration maps the confidence into a scene-invariant range, enabling a single threshold $\theta$ to remain effective across diverse tracking scenarios.
A higher nPSR value indicates a prominent and reliable peak, whereas a lower value suggests multiple competing responses, as illustrated in Fig.\ref{fig:npsr_visualization}.
When the nPSR falls below a predefined threshold, the tracker increases its reliance on historical trajectory information to maintain stable and accurate tracking, which will be detailed in the following part.

\subsubsection{Online Motion Model Refinement}

\begin{figure}[!t]
    \centering
    \includegraphics[width=0.95\linewidth]{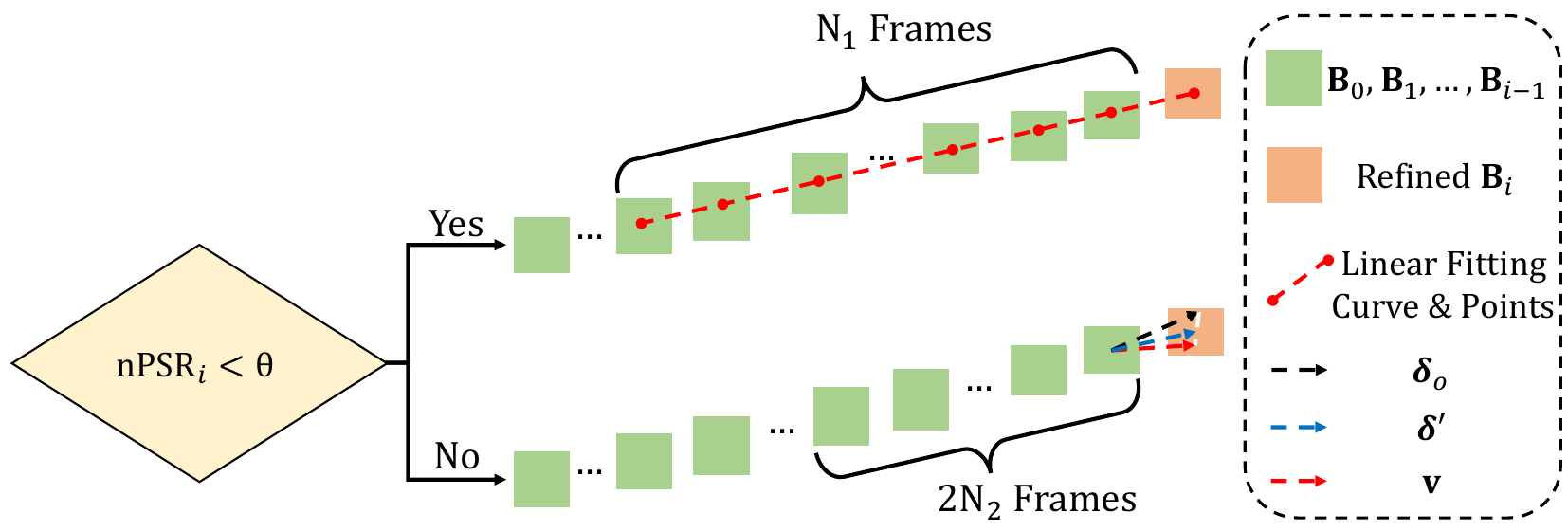}
    \caption{Illustration of the OMMR process.}
    \label{fig:motion_model_update}
    \vspace{-5pt}
\end{figure}

In standard online tracking, the bounding box prediction for the current frame relies solely on the response score map generated by the model, neglecting the potential value of historical trajectory information.
This can lead to issues such as tracking drift or failure, especially in the presence of occlusion, where the tracker may lose the target and subsequent predictions are updated based on erroneous results, causing error accumulation.
Considering that targets in satellite videos typically exhibit smooth and predictable motion trajectories, we leverage the motion vector derived from historical trajectories as prior information.
Accordingly, we propose the OMMR strategy, which adaptively adjusts the bounding box prediction for the current frame by integrating the nPSR value as a confidence measure, as illustrated in Fig.~\ref{fig:motion_model_update}.

Specifically, when the nPSR falls below the threshold $\theta$, we consider the response score map of the current frame to be unreliable, indicating that the model prediction may contain significant errors.
In this case, we rely entirely on long-term historical information for motion estimation.
By performing linear fitting on the historical trajectory information, we approximate the target's average velocity and scale variation.
Furthermore, we apply an exponential moving average (EMA) to smooth the predicted size, thereby obtaining the bounding box prediction for the current frame.
The detailed formulation is as follows:
\begin{equation}
\label{eq:motion_refine_average}
\begin{aligned}
(\mathbf{v}_c, \tilde{\mathbf{C}}_0) &= \varphi_{\text{fit}}(\mathbf{C}_0, \mathbf{C}_1, \dots, \mathbf{C}_{N_1 - 1}) \\
\mathbf{C}_{N_1} &= \varphi_{\text{val}}(\mathbf{v}_c, \tilde{\mathbf{C}}_0, N_1) \\
(\mathbf{v}_s, \tilde{\mathbf{S}}_0) &= \varphi_{\text{fit}}(\mathbf{S}_0, \mathbf{S}_1, \dots, \mathbf{S}_{N_1 - 1}) \\
\mathbf{S}_{N_1} &= \lambda\varphi_{\text{val}}(\mathbf{v}_s, \tilde{\mathbf{S}}_0, N_1) + (1-\lambda)\mathbf{S}_{N_1-1}
\end{aligned}
\end{equation}
where $\mathbf{C}_k$ and $\mathbf{S}_k$ represent the historical center coordinates and size of the bounding box in the $k$-th frame of $\mathcal{C}_{hist}$ and $\mathcal{S}_{hist}$, respectively.
$\varphi_{\text{fit}}$ and $\varphi_{\text{val}}$ denotes the linear fitting operation, and $\lambda$ is a hyperparameter of the EMA.
When the nPSR is high, the response score map of the current frame is considered reliable.
In this case, we only use the short-term instantaneous velocity to refine the center coordinates, while the width and height are updated directly using the EMA of the model's predicted results.
Calculation of instantaneous velocity is defined as follows:
\begin{equation}
\label{eq:motion_refine_instantaneous}
\mathbf{v} = \frac{1}{N_2} \sum_{k=0}^{N_2-1} \frac{\mathbf{C}_{k+N_2} - \mathbf{C}_k}{N_2}
\end{equation}
where the definition of $\mathbf{C}_k$ is consistent with the previous context.
Note that $N_1$ and $N_2$ denote the window sizes for long-term and short-term frames, respectively.
Given that most videos are sampled at 25 fps and the motion of targets in satellite videos is relatively slow, we typically set $N_1=50$, which means that the average velocity is estimated using the historical information from about past 2 seconds, and $N_2=10$, which corresponds to the instantaneous velocity.
Considering the varying nPSR ranges across different scenarios, we set the threshold to $\theta=0.5$, which effectively captures the motion trend of the target without introducing cumulative errors.
Note that no fitting is performed during the first $N_1$ frames of the tracking process for robustness.
The detailed steps of this approach are presented in Algorithm~\ref{alg:motion_refinement}.
It is noted that unlike the single Markov chain of Kalman Filter (KF), OMMR seamlessly decouples the timeline into dual-frequency temporal perceptions (the stable long-term $N_1$ window and the instantaneous short-term $N_2$ window).
This allows OMMR to instantly cut off erroneous visual updates during severe occlusions and fall back to the purely physical kinematic prior, fundamentally suppressing the cumulative errors that plague traditional KF.

\begin{algorithm}[!t]
\caption{Online Motion Model Refinement}\label{alg:motion_refinement}
    \begin{algorithmic}[1]
    \REQUIRE \hfill \\
    Frame window hyperparameters $[N_1, N_2]$ ($N_1 > 2N_2$); \\
    Score threshold $\theta$ for nPSR; \\
    Current frame index $i$; \\
    Response score map $S_i$; \\
    Historical box deque $\mathcal{B}_{hist}$ with capacity $N_1$;
    \ENSURE \hfill \\
    Refined bounding box $\mathbf{B}_i$.
    \STATE Calculate $\text{PSR}_i$ from $S_i$ using Eq.~(\ref{eq:psr});
    \STATE $\text{PSR}_{\text{max}} \gets \max(\text{PSR}_{\text{max}}, \text{PSR}_i)$;
    \STATE Calculate $\text{nPSR}_i$ using Eq.~(\ref{eq:npsr});
    \STATE Calculate model output $\mathbf{B}_o \gets \text{BBox}(\mathbf{C}_o, \mathbf{S}_o)$;
    \IF{$i > N_1$}
        \IF{$\text{nPSR}_i < \theta$}
            \STATE \emph{// Low confidence: Average Velocity}
            \STATE Extract $\mathcal{C}_{hist}$, $\mathcal{S}_{hist}$ from last $N_1$ frames in $\mathcal{B}_{hist}$;
            \STATE Calculate $\mathbf{C}_{N_1}$ and $\mathbf{S}_{N_1}$ based on Eq.~(\ref{eq:motion_refine_average});
            \STATE $\mathbf{B}_i \gets \text{BBox}(\mathbf{C}_{N_1},\mathbf{S}_{N_1})$;
        \ELSE
            \STATE \emph{// High confidence: Instantaneous Velocity}
            \STATE Extract $\mathcal{C}_{hist}$ from last $2N_2$ frames in $\mathcal{B}_{hist}$;
            \STATE Calculate motion vector $\mathbf{v}$ based on Eq.~(\ref{eq:motion_refine_instantaneous});
            \STATE $\alpha \gets (\text{nPSR}_i)^2$;
            \STATE $\boldsymbol{\delta}_o \gets \mathbf{C}_o - \mathbf{C}_{N_1-1}$;
            \STATE $\boldsymbol{\delta}' \gets \alpha \cdot \boldsymbol{\delta}_o + (1-\alpha) \cdot \mathbf{v}$;
            \STATE $\mathbf{B}_i \gets \text{BBox}(\mathbf{C}_{N_1-1} + \boldsymbol{\delta}', \lambda\mathbf{S}_o + (1-\lambda)\mathbf{S}_{N_1-1})$;
        \ENDIF
    \ELSE
        \STATE $\mathbf{B}_i \gets$ Standard prediction and update operation;
    \ENDIF
    \STATE Update $\mathcal{B}_{hist}$ by appending $\mathbf{B}_i$;
    \RETURN $\mathbf{B}_i$
    \end{algorithmic}
\end{algorithm}

\subsection{Training and Tracking Phase}

Our SiamGM adopts an anchor-free structure with three sub-branches, requiring the simultaneous optimization of classification, regression, and centerness losses.
Specifically, the classification loss employs our proposed CJCL, the regression loss uses the weighted IoU loss, and the centerness loss adopts the cross-entropy loss.
The loss functions are defined as follows:
\begin{equation}
\label{eq:loss_reg}
\mathcal{L}_{reg} = -\frac{\sum_{i=1}^{N_{pos}} c_i \log\frac{A_I + 1}{A_U + 1}}{\sum_{i=1}^{N_{pos}} c_i}
\end{equation}
\begin{equation}
\label{eq:loss_cen}
\mathcal{L}_{cen} = -\frac{1}{N_{pos}} \sum_{i=1}^{N_{pos}} \left[c_i\log(\hat{c}_i) + (1-c_i)\log(1-\hat{c}_i)\right]
\end{equation}
where $A_I$ and $A_U$ denote the intersection and union areas between the predicted box and the ground truth, respectively.
$c_i$ represents the centerness ground truth for the $i$-th positive sample, and $\hat{c}_i$ is the corresponding predicted value.
The overall loss function is defined as:
\begin{equation}
\label{eq:loss_total}
\mathcal{L}_{total} = \lambda_{cls}\mathcal{L}_{CJCL} + \lambda_{reg}\mathcal{L}_{reg} + \lambda_{cen}\mathcal{L}_{cen}
\end{equation}
where $\lambda_{cls}$, $\lambda_{reg}$, and $\lambda_{cen}$ are the weighting hyperparameters for the classification loss, regression loss, and centerness loss, respectively.
During training, we set $\lambda_{cls}=1$, $\lambda_{reg}=2$, and $\lambda_{cen}=1$ to balance the contributions of each loss component.

During the tracking phase, the model first extracts features using the backbone and neck, and then obtains the target response through depth-wise cross-correlation.
The response score map $S_i$ is generated based on the predictions from the three branches.
The initial model output $\mathbf{B}_{o}$ is determined by the peak position $t_i$ of $S_i$, for detailed calculation, please refer to~\cite{guo2020siamcar}.
This output is further refined by the OMMR strategy to produce the final tracking result $\mathbf{B}_i$.

%% file: experiment.tex
\section{Experiment}
\label{sec:experiment}

\subsection{Experimental Setup}

\subsubsection{Implementation Details}

Our experiments were conducted on an Ubuntu 22.04 platform equipped with a 24GB NVIDIA GeForce RTX 4090 GPU.
All training procedures were implemented using Python 3.11, CUDA 11.8, and PyTorch 2.0.1.
All of our models were trained for 20 epochs with a batch size of 48, using the stochastic gradient descent (SGD) optimizer with an initial learning rate of 0.005, a momentum of 0.9, and a weight decay of 0.0001.
To accelerate training, we adopt the modified ResNet-50~\cite{he2016deep} pre-trained on ImageNet from SiamRPN++~\cite{li2019siamrpn++} as the backbone.
During the first 10 epochs, the backbone weights are frozen to preserve the pre-trained representations.
In the subsequent 10 epochs, the last three convolutional layers of ResNet-50 are unfrozen for fine-tuning.
Considering the unique characteristics of satellite video tracking, we train our model exclusively on the VISO dataset with 30,000 image pairs per epoch.
Compared to general object tracking datasets, this setup significantly reduces the number of training pairs required while introducing minimal performance degradation.
The template and search images are resized to $127\times127$ and $255\times255$ pixels, respectively, and the data augmentation strategy follows that of the baseline.

\subsubsection{Testing Dataset}

We evaluate our model on two large-scale public benchmark datasets for satellite video single object tracking, namely SatSOT and SV248S, to validate the effectiveness and generalization ability of our method and compare with other state-of-the-art trackers.

\textit{SatSOT: }SatSOT~\cite{zhao2022satsot} consists of videos sourced from three commercial satellites: Jilin-1, Skybox, and Carbonite-2, with a frame rate of 10 or 25 fps and a duration of approximately 30 seconds.
The dataset contains 105 sequences covering four categories: trains, cars, airplanes, and ships. 
It introduces 11 challenging attributes, such as background clutter (BC), tiny object (TO), and partial/full occlusion (POC/FOC), which best reflect the characteristics of satellite video.
Notably, more than 70\% of the sequences contain targets with an area smaller than 1,000 pixels, and even more than 30\% contain targets with an area smaller than 100 pixels.

\textit{SV248S: }SV248S~\cite{li2022deep} consists of videos captured by the Jilin-1 video satellite at a frame rate of 25 fps, with durations ranging from 20 to 30.12 seconds.
The dataset comprises 248 satellite video sequences across six scenes, with approximately 40 targets per scene, covering four categories, including ships, vehicles, large vehicles, and airplanes.
It introduces 3 basic properties, 3 frame state flags, and 10 sequence attributes to comprehensively characterize SVs in the SV248S dataset.
In SV248S, object size (OSs) is defined as the diagonal length of the ground-truth bounding box. 
The average OSs for the four target categories are 15.2, 8.1, 15.6, and 47.8 pixels, respectively.

\subsubsection{Evaluation Metrics}

We adopt the one-pass evaluation (OPE) protocol~\cite{wu2015object} to assess tracking performance, where the target state is initialized in the first frame and all subsequent frames are predicted solely by the tracker.
Three metrics are reported: precision, norm precision and success.
It is important to note that all three metrics are averaged cross video sequences in the dataset.

\textit{Precision: }The precision plot measures the percentage of frames in which the estimated center location falls within different distance thresholds from the ground-truth position, referred to as the center location error (CLE) threshold.
Since most targets in satellite videos are relatively small, we report the precision score at a CLE threshold of 5 pixels following~\cite{zhao2022satsot}.
Additionally, given that many existing trackers still follow the traditional OTB benchmark, we also report the precision score at a CLE threshold of 20 pixels to facilitate comprehensive comparisons with other trackers.
Note that the precision reported in all subsequent ablation studies defaults to the 5-pixel threshold.

\textit{Norm Precision: }To evaluate tracking performance across targets of varying scales and aspect ratios in ablation studies, we also plot norm precision under different normalized CLE thresholds, where the CLE is normalized by the size of the ground-truth bounding box.
We report the norm precision rate at a normalized CLE threshold of 0.5, which is more tolerant and better reflects the tracking performance of targets in satellite videos.

\textit{Success: }The success plot measures the percentage of frames in which the IoU between the estimated and ground-truth bounding boxes exceeds different overlap thresholds, and we report the area under curve (AUC) of the success plot as the success rate.

\begin{table*}[!t]
    \caption{Overall Performance on SatSOT and SV248S. HOG is Histogram of Oriented Gradients, CN is Color Names, CF is Convolutional Feature, and TF is Transformer Feature. The P-5. is Precision Rate with CLE Threshold of 5 Pixels, P-20. of 20 Pixels, and the S. is Success Rate. The \textcolor{red}{\textbf{Red}} Represents the Best Performance, \textcolor{blue}{\textbf{Blue}} Represents the Second-place, and \textcolor{green}{\textbf{Green}} Represents Third-place. The values in parentheses indicate the absolute performance gains over the baseline SiamCAR.}
    \label{tab:overall_performance}
    \centering
    \begin{tabular}{lcccccccc}
    \toprule
    \multirow{2.5}{*}{\textbf{Trackers}} & \multirow{2.5}{*}{\textbf{Source}} & \multirow{2.5}{*}{\textbf{Features}} & \multicolumn{3}{c}{\textbf{SatSOT}} & \multicolumn{3}{c}{\textbf{SV248S}} \\
    \cmidrule{4-6} \cmidrule{7-9}
    &  &  & \textbf{P-5. (\%)} & \textbf{P-20. (\%)} & \textbf{S. (\%)} & \textbf{P-5. (\%)} & \textbf{P-20. (\%)} & \textbf{S. (\%)} \\
    \midrule
    KCF         & TPAMI2014     & HOG    & 55.3 & 62.2 & 40.0 & 73.1 & 76.7 & 46.2 \\
    ECO-HC      & CVPR2017      & HOG+CN & 58.6 & 66.4 & 43.8 & 71.1 & 77.2 & 44.0 \\
    \midrule
    SiamCAR     & CVPR2020      & CF     & 60.0 & 67.8 & 42.7 & 71.8 & 77.1 & 41.4 \\
    Mixformer-L & CVPR2022      & TF     & 46.7 & 52.8 & 37.8 & 33.9 & 37.4 & 14.9 \\
    SeqTrack    & CVPR2023      & CF+TF  & 52.3 & 56.6 & 40.1 & 41.7 & 43.4 & 16.6 \\
    ARTrack     & CVPR2023      & CF+TF  & 58.4 & 64.8 & 44.3 & 47.0 & 55.6 & 19.9 \\
    SmallTrack  & TGRS2023      & CF     & 53.0 & 62.0 & 36.1 & 46.5 & 53.7 & 18.1 \\
    SiamMDM     & TGRS2023      & CF     & 61.1 & 68.0 & 46.1 & 71.6 & 77.0 & 45.1 \\
    SiamTM      & RS2023        & CF     & 60.8 & 69.0 & 47.5 & 75.3 & 80.0 & 48.7 \\
    AQATrack    & CVPR2024      & CF+TF  & 51.8 & 61.8 & 39.6 & 37.8 & 52.8 & 16.5 \\
    HIPTrack    & CVPR2024      & CF+TF  & 54.1 & 61.9 & 41.9 & 33.3 & 36.7 & 15.4 \\
    SVLPNet     & TCSVT2024     & CF+TF  & 58.4 & 67.2 & 46.5 & 76.9 & 82.3 & 47.1 \\
    TATrans     & TGRS2024      & CF+TF  & 57.6 & 67.5 & 45.6 & 61.8 & 71.4 & 33.9 \\
    Adaptive MT & TGRS2024      & CF+TF  & 60.5 & 70.3 & 42.4 & \textcolor{green}{\textbf{77.8}} & \textcolor{green}{\textbf{84.6}} & 43.7 \\
    ORTrack     & CVPR2025      & TF     & 46.0 & 57.0 & 38.0 & 38.2 & 46.5 & 17.5 \\ 
    SGLATrack   & CVPR2025      & TF     & 48.3 & 58.9 & 39.1 & 40.9 & 49.0 & 18.6 \\
    LMTrack     & AAAI2025      & TF     & 56.9 & 70.5 & 47.0 & 69.0 & 81.6 & 33.2 \\
    EHTracker   & JSTARS2025    & CF     & \textcolor{blue}{\textbf{62.2}} & 70.2 & 47.2 & 74.3 & 79.5 & \textcolor{green}{\textbf{49.0}} \\
    SiamS\textsuperscript{2}F & TGRS2025 & CF   & \textcolor{green}{\textbf{62.1}} & \textcolor{green}{\textbf{71.0}} & \textcolor{blue}{\textbf{48.6}} & 76.3 & 81.5 & \textcolor{blue}{\textbf{50.6}} \\
    TSTrans     & TGRS2026      & CF+TF  & 61.4 & \textcolor{blue}{\textbf{73.0}} & \textcolor{red}{\textbf{53.2}} & \textcolor{blue}{\textbf{80.6}} & \textcolor{blue}{\textbf{89.0}} & 40.8 \\
    SiamGM      & Ours          & CF     & \textcolor{red}{\textbf{65.9}}\textbf{($\uparrow$5.9)} & \textcolor{red}{\textbf{74.2}}\textbf{($\uparrow$6.4)} & \textcolor{green}{\textbf{47.7}}\textbf{($\uparrow$5.0)} & \textcolor{red}{\textbf{85.2}}\textbf{($\uparrow$13.4)} & \textcolor{red}{\textbf{89.3}}\textbf{($\uparrow$12.2)} & \textcolor{red}{\textbf{50.9}}\textbf{($\uparrow$9.5)} \\
    \bottomrule
    \end{tabular}
    \vspace{-5pt}
\end{table*}

\subsection{Comparison With State-of-the-Art Trackers}

To comprehensively evaluate the performance of our SiamGM, we compare it with 20 representative state-of-the-art trackers.
These trackers can be categorized into three groups based on feature extraction mechanisms:
1) Hand-crafted feature-based trackers: KCF~\cite{henriques2014high} and ECO-HC~\cite{danelljan2017eco}, which rely on HOG and Color Names features.
2) CNN-based trackers: SiamCAR~\cite{guo2020siamcar}, SmallTrack~\cite{xue2023smalltrack}, SiamMDM~\cite{yang2023siammdm}, SiamTM~\cite{yang2023single}, EHTracker~\cite{yang2024ehtracker}, and SiamS\textsuperscript{2}F~\cite{lv2025siams2f}, which utilize deep convolutional features and Siamese network architectures.
3) Transformer-based and Hybrid trackers: Mixformer-L~\cite{cui2022mixformer}, SeqTrack~\cite{chen2023seqtrack}, ARTrack~\cite{wei2023autoregressive}, AQATrack~\cite{xie2024autoregressive}, HIPTrack~\cite{cai2024hiptrack}, SVLPNet~\cite{wang2024satellite}, Adaptive MT~\cite{zhang2024adaptive}, TATrans~\cite{lai2023target}, ORTrack~\cite{wu2025learning}, SGLATrack~\cite{xue2025similarity}, LMTrack~\cite{xu2025less}, and TSTrans~\cite{li2026tstrans} which leverage the attention mechanism of Transformers or a combination of CNN and Transformer features to capture long-range dependencies.

\begin{figure}[!t]
    \centering
    \includegraphics[width=0.95\linewidth]{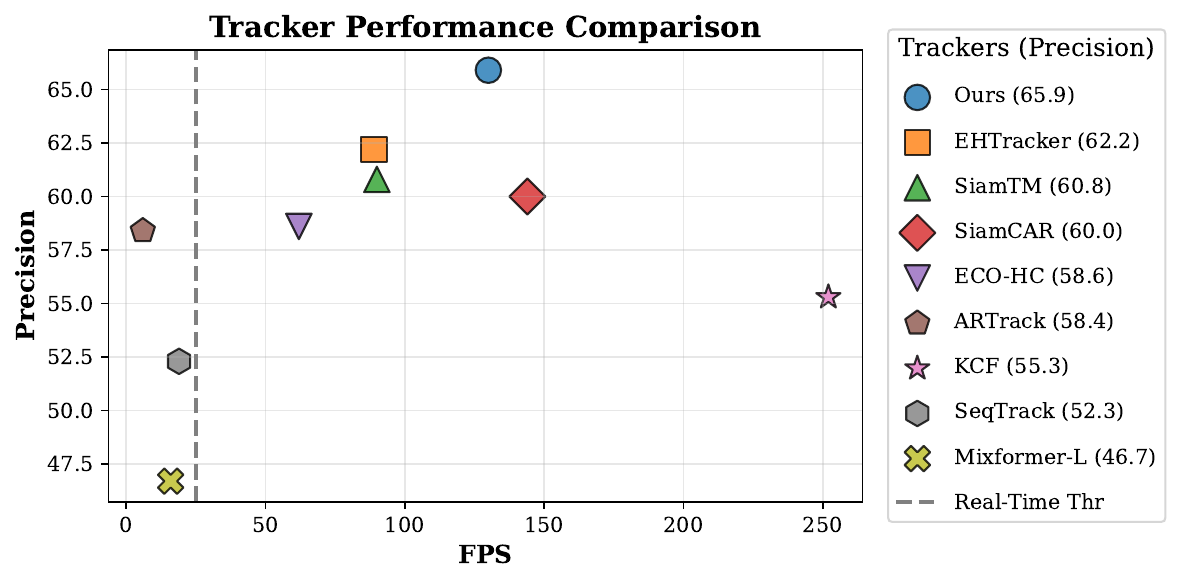}
    \caption{Precision vs. FPS Comparison on SatSOT benchmark. The horizontal axis represents the tracking speed, while the vertical axis represents the precision score.}
    \label{fig:precision_fps}
    \vspace{-5pt}
\end{figure}

The overall quantitative tracking results on the benchmark datasets are extensively summarized in Table~\ref{tab:overall_performance}.
On the challenging SatSOT dataset, our proposed SiamGM demonstrates outstanding generalization and robustness, outperforming all other compared trackers in both precision metrics, while simultaneously securing highly competitive results in success rate.
Specifically, it achieves an impressive precision of 65.9\% at a 5-pixel threshold (P-5) and 74.2\% at a 20-pixel threshold (P-20).
This not only surpasses the recent state-of-the-art tracker TSTrans by substantial absolute margins of 4.5\% and 1.2\%, respectively, but also yields significant absolute improvements of 5.9\% and 6.4\% over the baseline.
In terms of the success rate, SiamGM secures a highly competitive 47.7\%, bringing a 5.0\% absolute increase compared to the baseline.
Similarly, on the SV248S dataset, our proposed SiamGM maintains its robust tracking capability and achieves highly competitive performance.
Specifically, it achieves P-5, P-20, and success rates of 85.2\%, 89.3\%, and 50.9\%, respectively, representing absolute improvements of 13.4\%, 12.2\%, and 9.5\% over the baseline.
Compared to other advanced trackers like SiamS\textsuperscript{2}F and TSTrans, SiamGM exhibits distinct advantages or comparable results both in precision and success metrics, further consistently proving its exceptional generalization and spatial-temporal adaptability across different satellite sensors and target categories.
More importantly, our model achieves this accuracy surge without incurring heavy parameter burdens or computational overhead.
Specifically, the TAM module is constructed as a lightweight module, the GCLA method introduces virtually zero extra computations, and the calculations required for OMMR are negligible.
Overall, the inference speed is slightly reduced compared to the baseline, remaining well above the standard requirement for real-time tracking.
This remarkable efficiency fundamentally benefits from the anchor-free mechanism and the elegant fully convolutional architecture of SiamGM.
Further comprehensive comparisons regarding evaluation metrics versus tracking speeds are explicitly illustrated in Fig.~\ref{fig:precision_fps}.

\begin{table}[!t]
    \caption{Comprehensive Ablation Studies on SatSOT. \textbf{Bold} Highlights the Best Results, and \underline{Underline} Marks the Second-place Results. All three metrics are reported as percentages.}
    \label{tab:module_ablation}
    \centering
    \resizebox{\columnwidth}{!}{%
    \begin{tabular}{l|ccc|cccc}
    \toprule
    \textbf{Method} & \textbf{TAM} & \textbf{GCLA} & \textbf{OMMR} & \textbf{P.} & \textbf{N.P.} & \textbf{S.} & \textbf{FPS} \\
    \midrule
    SiamCAR      & $\times$ & $\times$ & $\times$             & 60.0 & 70.8 & 42.7 & 142 \\
    SiamCAR-TAM  & \checkmark & $\times$ & $\times$           & 61.9 & 73.3 & 44.2 & 136 \\
    SiamCAR-GCLA & $\times$ & \checkmark & $\times$           & 61.1 & 71.9 & 43.5 & 142 \\
    SiamCAR-M    & $\times$ & $\times$ & \checkmark           & \underline{63.3} & 74.1 & 44.6 & 132 \\
    SiamCAR-G    & \checkmark & \checkmark & $\times$         & 63.1 & \underline{74.2} & \underline{45.2} & 136 \\
    SiamGM       & \checkmark & \checkmark & \checkmark       & \textbf{65.9} & \textbf{76.6} & \textbf{47.7} & 130 \\
    \bottomrule
    \end{tabular}}
    \vspace{-5pt}
\end{table}

Furthermore, we observe that modern heavy Transformer-based general trackers (such as Mixformer-L and ORTrack) suffer significant performance degradation in satellite scenarios.
This vulnerability primarily stems from their strong reliance on rich appearance and texture details, which are conspicuously absent for tiny satellite targets.
This severe domain gap further underscores the necessity and superiority of customized architectures like SiamGM, which elegantly exploit structural topologies and motion prior instead of purely relying on appearance features.

\subsection{Ablation Study}

To validate the effectiveness of the proposed modules, we conduct extensive ablation studies on the SatSOT dataset, with the detailed results presented in Table~\ref{tab:module_ablation}.
Overall, SiamGM systematically optimizes the baseline model from two complementary perspectives: geometry-topology and motion, corresponding to the SiamCAR-G and SiamCAR-M variants.
Furthermore, since SiamCAR-G integrates both the TAM and the GCLA method, we decouple it into SiamCAR-TAM and SiamCAR-GCLA for a more granular analysis.
Specifically, compared to the baseline, SiamCAR-TAM yields a 2.5\% absolute improvement in normalized precision.
However, its gains in precision and success rate are relatively marginal, primarily because TAM focuses on capturing topological correspondences rather than directly refining the bounding box coordinates for extreme tiny targets.
Conversely, SiamCAR-GCLA secures a 1.1\% absolute gain in precision, which effectively bridges the localization gap for targets with large aspect ratios, though its isolated impact on regular tiny targets remains limited.
Notably, when seamlessly integrating these two components, SiamCAR-G achieves synergistic absolute improvements of 3.1\%, 3.4\%, and 2.5\% in precision, normalized precision, and success rate, respectively.
This demonstrates a strong complementarity between TAM and GCLA.
Furthermore, SiamCAR-M relies on temporal motion modeling to significantly boost tracking robustness, yielding absolute improvements of 3.3\%, 3.3\%, and 1.9\% across the three metrics.
This enhancement is especially pronounced in severe occlusion scenarios, profoundly alleviating tracking drift and tiny target loss.
Ultimately, our comprehensive SiamGM framework delivers remarkable absolute surges of 5.9\%, 5.8\%, and 5.0\% in precision, normalized precision, and success over the baseline.
This comprehensively validates the individual efficiency of each module and the powerful complementarity of the geometric-topological perception and motion prior paradigms.

\begin{table}[t]
    \caption{Ablation Study on $\gamma$ Parameter in the GCLA Method on SatSOT.}
    \label{tab:la_ablation}
    \centering
    \begin{tabular}{c|ccc}
    \toprule
    $\boldsymbol{\gamma}$ & \textbf{P. (\%)} & \textbf{N.P. (\%)} & \textbf{S. (\%)} \\
    \midrule
    0.25 & 60.5 & 71.4 & 43.0 \\
    0.5  & \textbf{61.1} & \textbf{71.9} & \textbf{43.5} \\
    0.75 & 59.8 & 71.5 & 42.9 \\
    1.0  & 60.2 & 71.1 & 42.5 \\
    \bottomrule
    \end{tabular}
    \vspace{-5pt}
\end{table}

\begin{table}[!t]
    \caption{Ablation Study on the Gating Confidence of OMMR on SatSOT.}
    \label{tab:motion_metric_ablation}
    \centering
    \begin{tabular}{c|cccc}
    \toprule
    \textbf{Metric} & \textbf{P. (\%)} & \textbf{N.P. (\%)} & \textbf{S. (\%)} & \textbf{FPS}\\
    \midrule
    Reliability Coeff & 63.1 & 73.7 & 45.9 & 128 \\
    Location Map      & 63.0 & 73.6 & 44.5 & 122 \\
    nPSR              & \textbf{65.9} & \textbf{76.6} & \textbf{47.7} & \textbf{130} \\
    \bottomrule
    \end{tabular}
    \vspace{-5pt}
\end{table}

To further investigate the specific impact of the hyperparameter $\gamma$ on the centerness distribution within the GCLA method, we conduct an additional ablation study.
As summarized in Table~\ref{tab:la_ablation}, the model achieves optimal performance across all metrics when $\gamma$ is set to 0.5.
An excessively large $\gamma$ causes the valid centerness scores to spread too broadly along the major axis, blurring the discrimination boundary between the target feature and the background.
Conversely, an overly small $\gamma$ sharply diminishes the intended shape-constraining effect, rendering the module practically ineffective, as demonstrated in Fig.~\ref{fig:centerness_visualization}\subref{fig:centerness_1d_multi_gamma}.

Additionally, we study the choice of the gating confidence signal within the OMMR framework.
We compare two widely adopted alternatives, the reliability coefficient~\cite{yang2023siammdm} and the location map metric~\cite{zhou2025siamtitp}, with the nPSR adopted in this work.
As detailed in Table~\ref{tab:motion_metric_ablation}, nPSR achieves the best accuracy across all metrics at the highest inference speed, confirming that an online self-calibrated, scene-invariant confidence is better suited as the gate of OMMR while introducing negligible overhead.

\begin{figure*}[!t]
    \centering
    \includegraphics[width=0.95\linewidth]{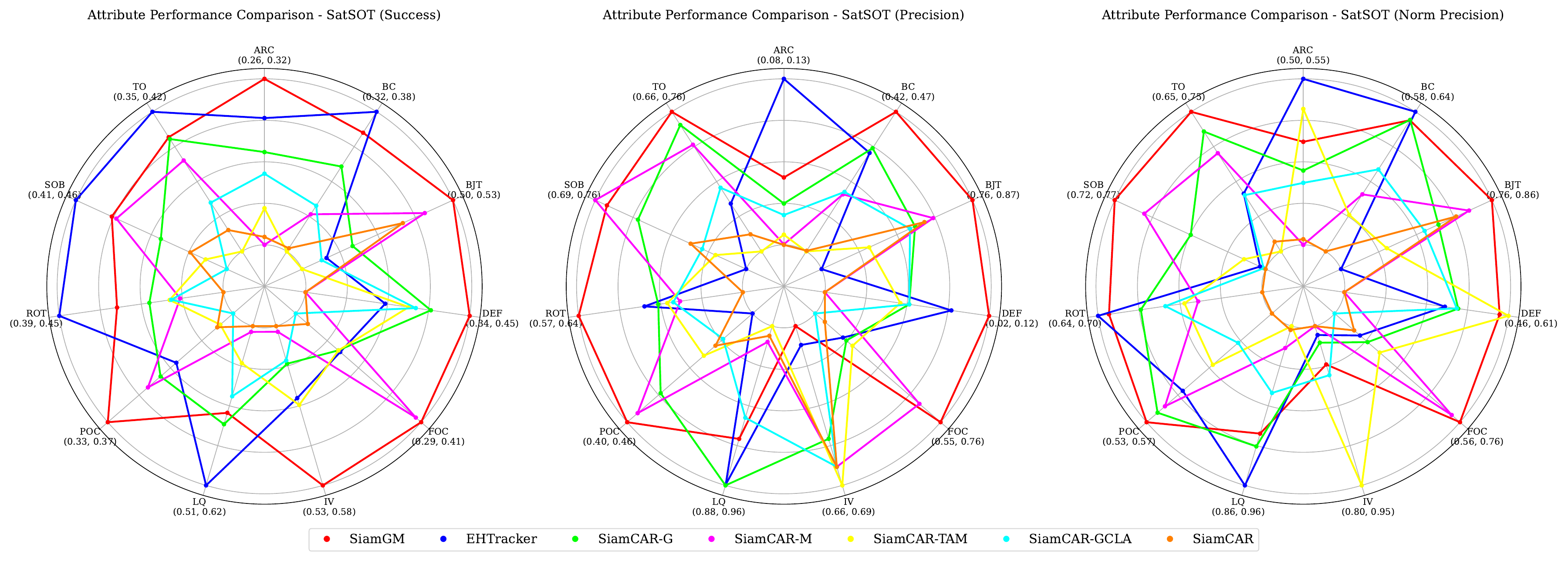}
    \caption{Attribute-based performance comparisons evaluating SiamGM, its ablated variants, and recent SOTA SNN-based tracker on the SatSOT dataset via radar charts. Our SiamGM (in red) exhibits distinct advantages and robust generalization in most challenging scenarios.}
    \label{fig:radar_attributes}
    \vspace{-5pt}
\end{figure*}

\begin{table}[!t]
    \centering
    \caption{Attributes for Performance Analysis in SatSOT.}
    \label{tab:attributes}
    \begin{tabular}{c|l|c}
        \toprule
        \textbf{Attribute} & \textbf{Description} & \textbf{Sequences} \\
        \midrule
        BC  & Background Clutter & 45 \\
        IV  & Illumination Variation & 3 \\
        LQ  & Low Quality & 13 \\
        ROT & Rotation & 56 \\
        POC & Partial Occlusion & 34 \\
        FOC & Full Occlusion & 12 \\
        TO  & Tiny Object & 21 \\
        SOB & Similar Object & 27 \\
        BJT & Background Jitter & 14 \\
        ARC & Aspect Ratio Change & 26 \\
        DEF & Deformation & 6 \\
        \bottomrule
    \end{tabular}
    \vspace{-5pt}
\end{table}

\begin{figure*}[!t]
    \centering
    \includegraphics[width=0.24\linewidth]{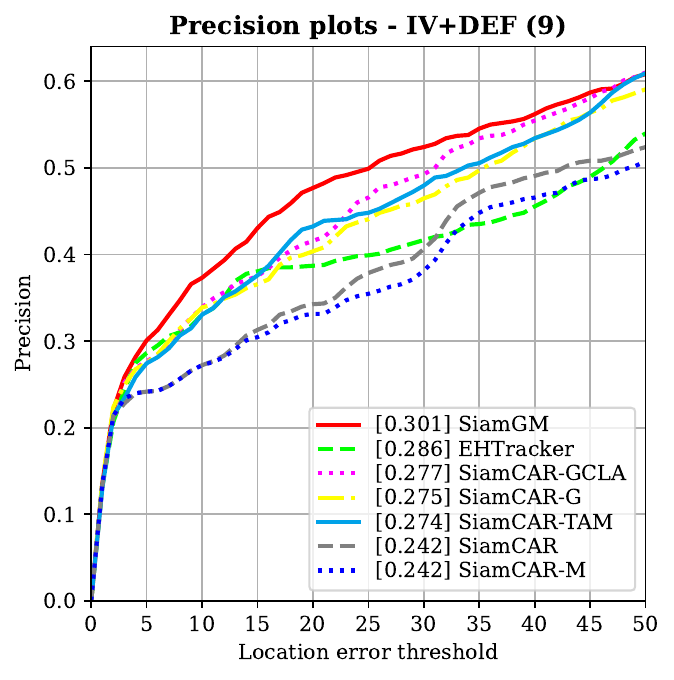}\hfill
    \includegraphics[width=0.24\linewidth]{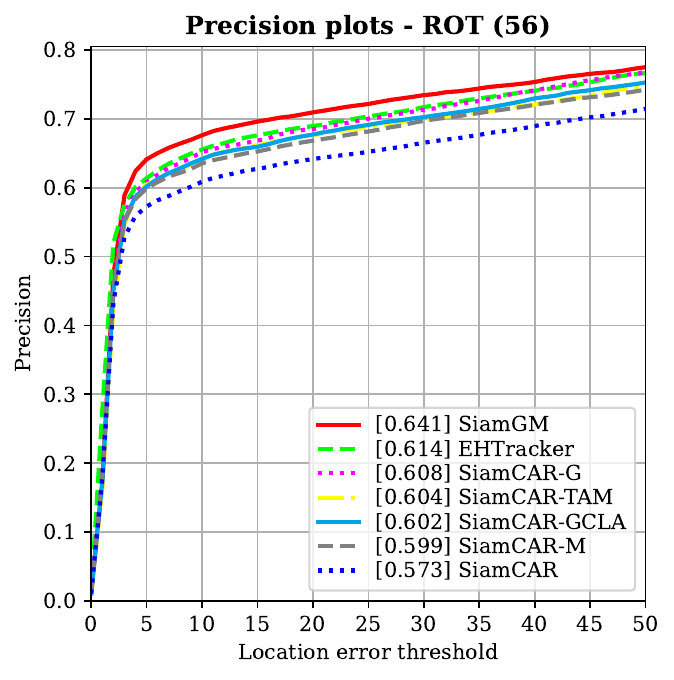}\hfill
    \includegraphics[width=0.24\linewidth]{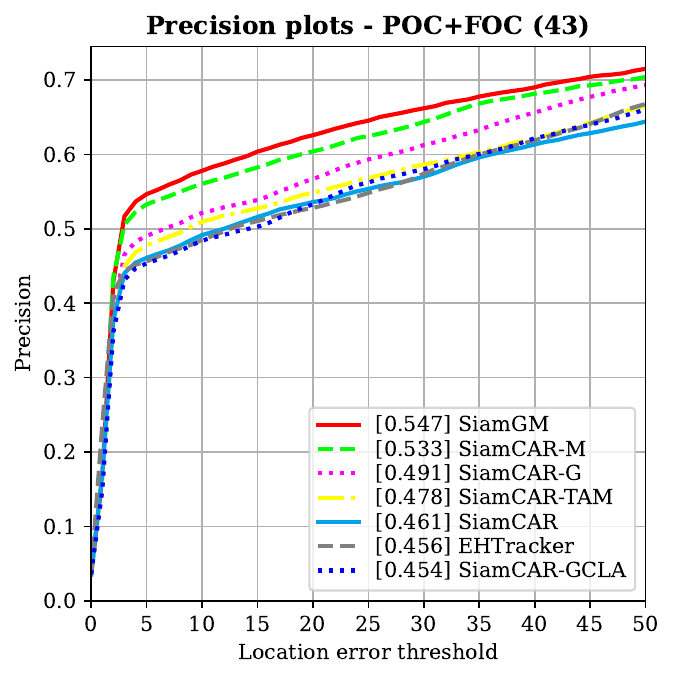}\hfill
    \includegraphics[width=0.24\linewidth]{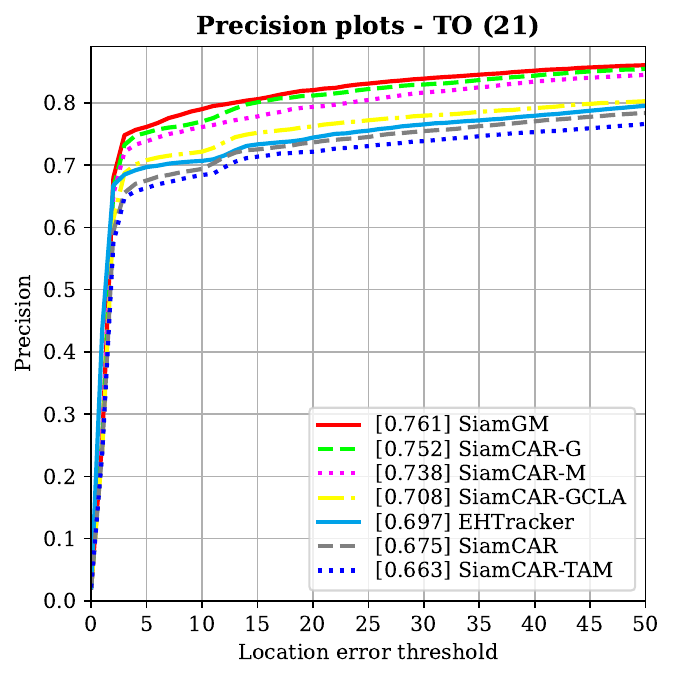}\\
    \includegraphics[width=0.24\linewidth]{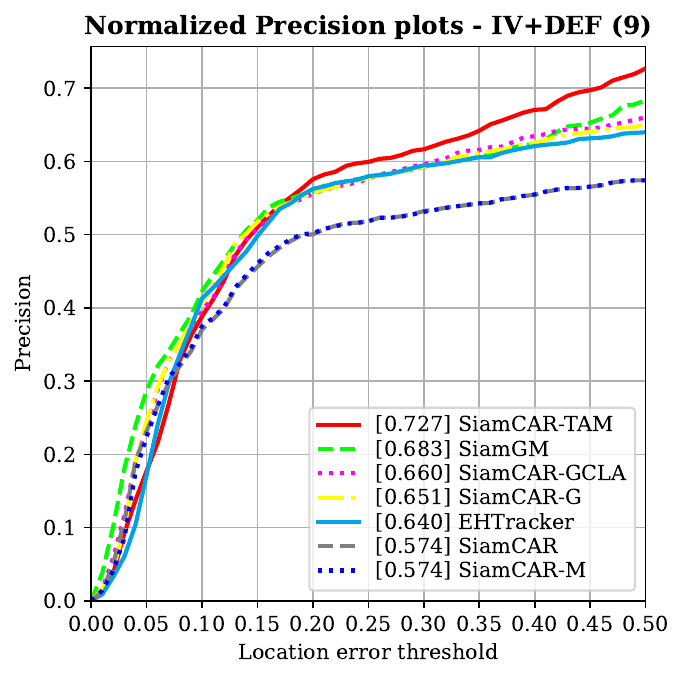}\hfill
    \includegraphics[width=0.24\linewidth]{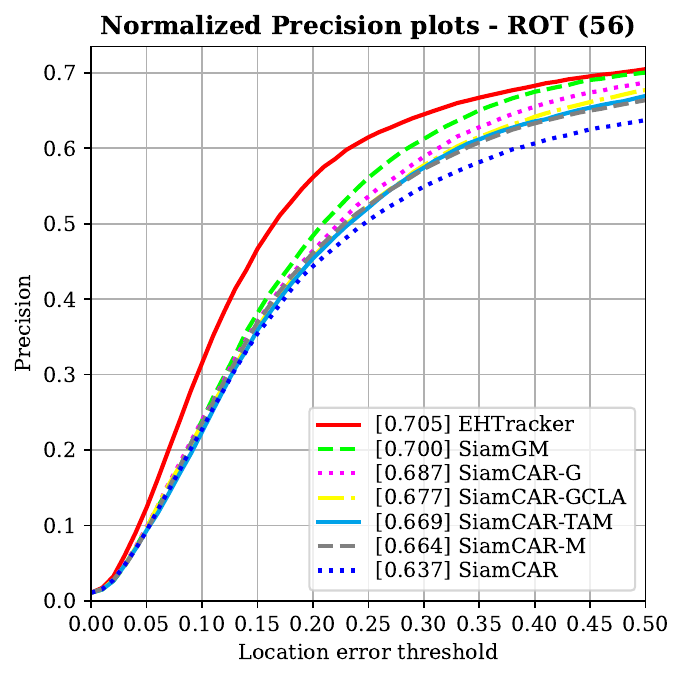}\hfill
    \includegraphics[width=0.24\linewidth]{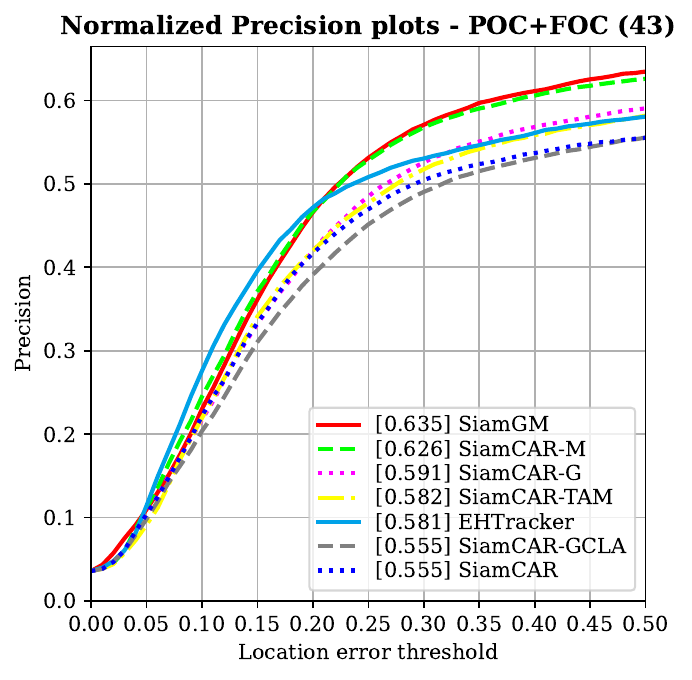}\hfill
    \includegraphics[width=0.24\linewidth]{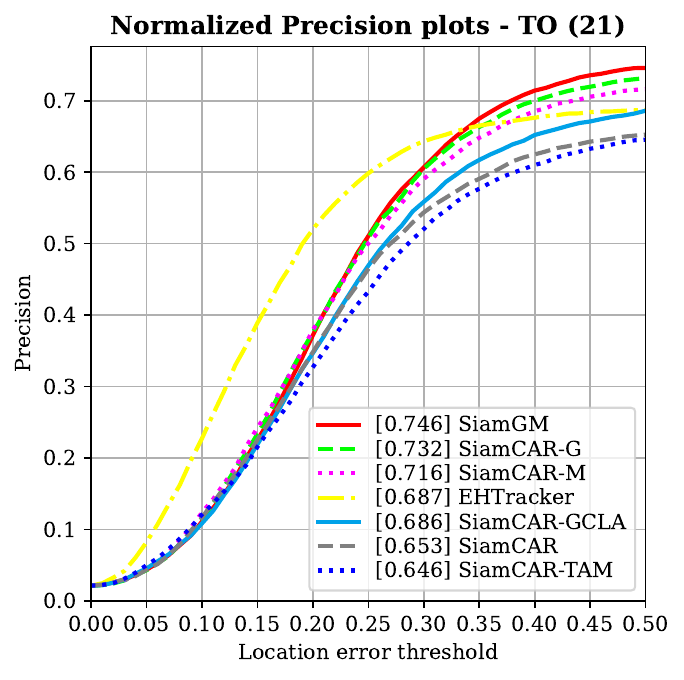}\\
    \includegraphics[width=0.24\linewidth]{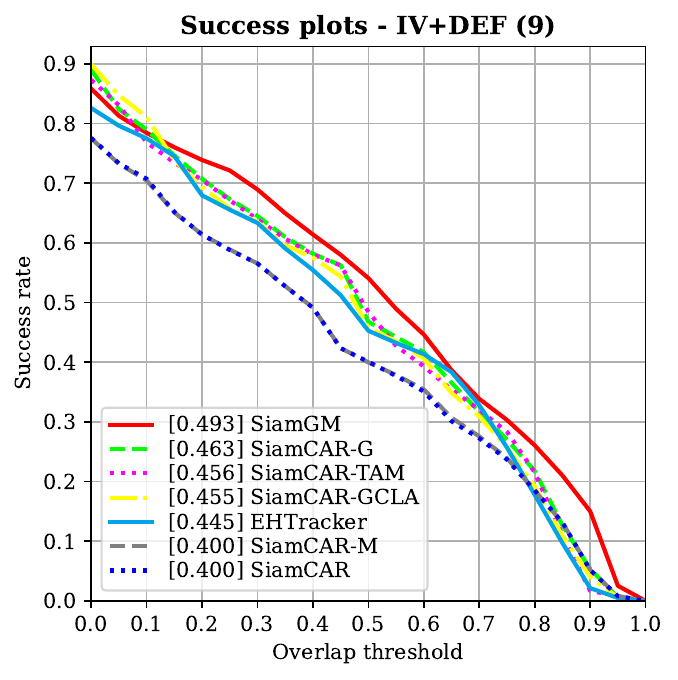}\hfill
    \includegraphics[width=0.24\linewidth]{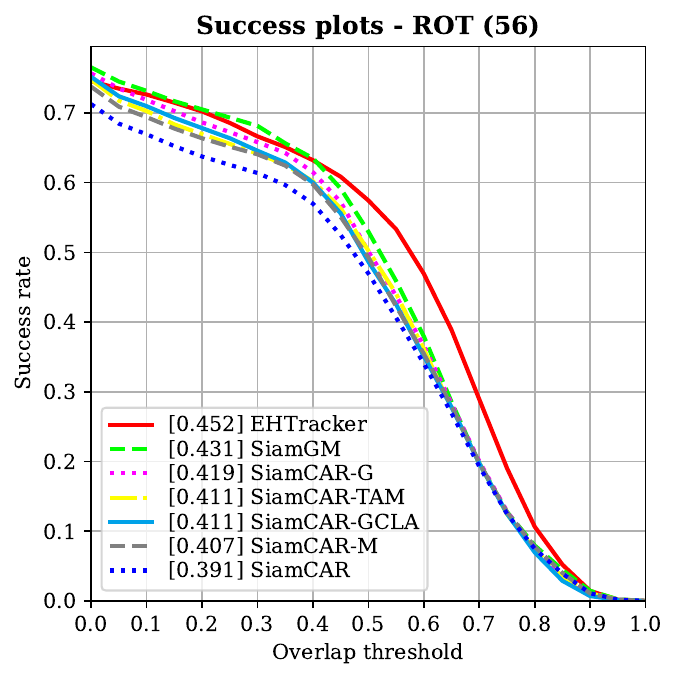}\hfill
    \includegraphics[width=0.24\linewidth]{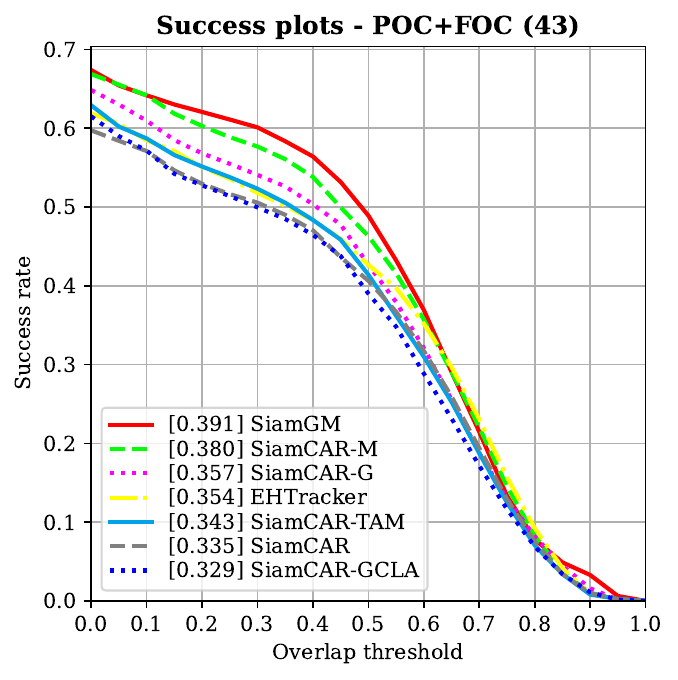}\hfill
    \includegraphics[width=0.24\linewidth]{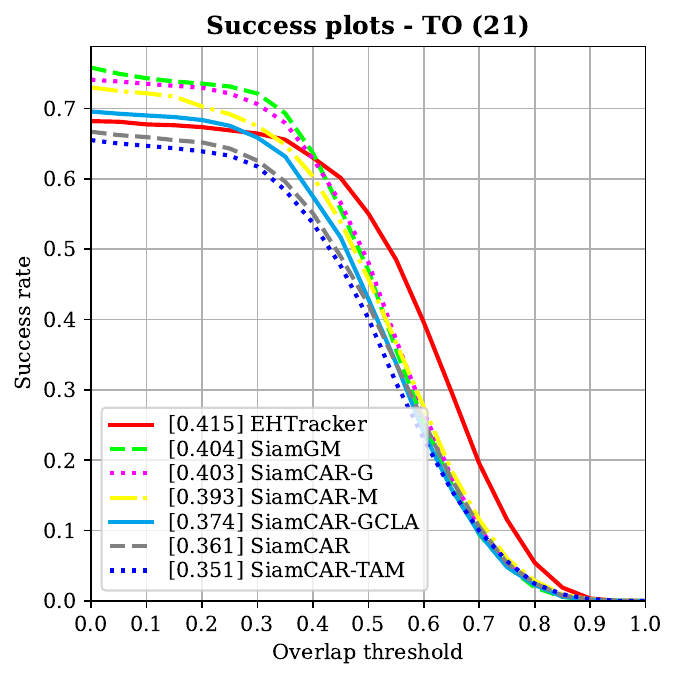}
    \caption{Attribute-based performance comparison of SiamGM, its ablated variants, and state-of-the-art tracker on the SatSOT dataset. The twelve sub-figures present precision, normalized precision, and success plots under four representative challenging scenarios (IV+DEF, ROT, POC+FOC, TO) including six attributes. SiamGM consistently demonstrates superior performance, especially under severe deformation, rotation, occlusion, and tiny object scenarios.}
    \label{fig:attr_performance}
    \vspace{-5pt}
\end{figure*}

%% file: discussion.tex
\section{Discussion}
\label{sec:discussion}

\subsection{Attribute-Based Analysis}

To comprehensively evaluate the effectiveness of SiamGM across diverse scenarios, we first analyze its performance on all 11 attributes in the SatSOT benchmark, with specific definitions and sequence statistics provided in Table~\ref{tab:attributes}.
To clearly visualize the evaluation metrics for each attribute, we present radar charts comparing SiamGM, its ablated variants, and state-of-the-art SNN-based EHTracker~\cite{yang2024ehtracker} on the SatSOT dataset, as shown in Fig.~\ref{fig:radar_attributes}.
These results demonstrate that SiamGM achieves superior generalization and distinct advantages over both the baseline SiamCAR and other trackers across nearly all challenging conditions.
Subsequently, to precisely dissect the individual contributions of our carefully designed modules, we conduct a more fine-grained evaluation focusing on six representative extreme attributes: IV, DEF, ROT, POC, FOC, and TO. By systematically comparing under these distinctly challenging conditions, we can isolate and verify the underlying mechanisms and effectiveness of each component.

\subsubsection{Effectiveness of TAM}

As analyzed in the methodology, the TAM is designed to capture fine-grained topological correspondences against intense appearance and spatial deformation.
The IV and DEF attributes precisely represent these challenges: in IV scenarios, target grayscale and texture feature averages can change drastically, while in DEF scenarios (especially for train sequences), targets exhibit severe non-rigid structural deformations along their trajectories.
Given the limited number of sequences for these two attributes, we merge them to ensure a statistically reliable evaluation.
As shown in the first column of Fig.~\ref{fig:attr_performance}, SiamGM achieves a substantial performance margin under these conditions.
Notably, this improvement primarily stems from the integration of the TAM within the SiamCAR-G variant, yielding absolute gains of 3.2\% vs. 5.9\%, 15.3\% vs. 10.9\%, and 5.6\% vs. 9.3\% in precision, normalized precision, and success rate, respectively (the paired values conventionally denote the respective improvements of the isolated module and the final SiamGM over the baseline hereafter).
This significant gain robustly validates the TAM's capability in maintaining target topological structures under complex visual transformations.

\subsubsection{Effectiveness of GCLA}

The core motivation behind our Geometry-Constrained Label Assignment strategy is to incorporate macroscopic shape prior into target localization.
Considering the nature of standard horizontal bounding box annotations in satellite datasets, object rotation inevitably leads to drastic and sudden fluctuations in the apparent aspect ratio. Consequently, ROT serves as an ideal attribute to validate this mechanism.
The performance curves in the second column of Fig.~\ref{fig:attr_performance} demonstrate a significant gain for SiamGM under the ROT attribute, which is predominantly attributed to the GCLA method within the SiamCAR-G variant, yielding absolute improvements of 2.9\% vs. 6.8\%, 4.0\% vs. 6.3\%, and 2.0\% vs. 4.0\% in precision, normalized precision, and success rate, respectively.
This substantial improvement validates its effectiveness in dynamically adapting to large aspect ratio variations and mitigating boundary degradation.

\subsubsection{Effectiveness of OMMR}

The OMMR strategy is specifically developed to refine trajectory predictions by exploiting physical motion prior, making it crucial when appearance-based tracking momentarily fails.
Such signal losses frequently occur during POC and FOC.
Furthermore, for tiny objects, the extremely limited spatial resolution can easily cause the predicted bounding box to completely detach from the target under minor disturbances.
Therefore, we select the combined POC+FOC and the TO attributes to assess the OMMR inference. The plots in the third and fourth columns of Fig.~\ref{fig:attr_performance} reveal that SiamGM outperforms other variants remarkably in these challenging situations.
The dominant source of this performance boost is the SiamCAR-M variant, strongly substantiating the OMMR's superior capacity to maintain tracking continuity by safely integrating historical motion vectors, yielding absolute gains of 7.2\% vs. 8.6\%, 7.1\% vs. 8.0\%, and 4.5\% vs. 5.6\% in precision, normalized precision, and success rate for POC+FOC, and 6.3\% vs. 8.6\%, 6.3\% vs. 9.3\%, and 3.2\% vs. 4.3\% for TO, respectively.

\begin{figure}[!t]
    \centering
    \includegraphics[width=0.95\linewidth]{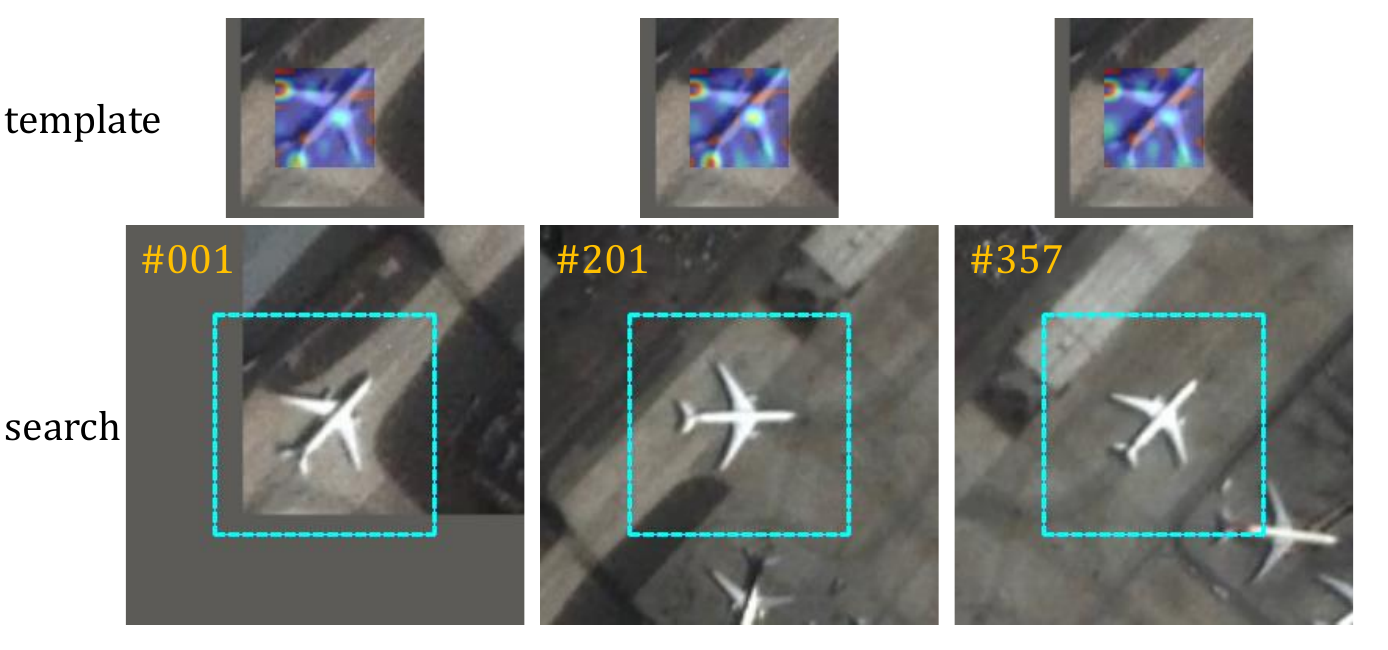}
    \caption{Visualization of the template saliency maps for the \texttt{plane\_02} sequence, which contains both IV and ROT attributes. The consistent focus on geometrically salient regions (e.g., wings and fuselage) across different orientations validates the effectiveness of TAM in enhancing geometric structural perception.}
    \label{fig:vis_ifga}
    \vspace{-5pt}
\end{figure}

\subsection{Visual Analysis}

To validate the geometric structural perception capability of TAM, we conduct a visualization analysis on the \texttt{plane\_02} sequence, which contains both IV and ROT attributes---exhibiting significant grayscale changes due to illumination and rotation variations.
As shown in Fig.~\ref{fig:vis_ifga}, the top row displays the template saliency maps, computed by summing the attention weights from the bounding box region in the search image to each template location, as defined by:
\begin{equation}
    \mathcal{S}_j = \sum_{i\in V_s} W_{s,t}^{(i,j)}
\end{equation}
where $W_{s,t}^{(i,j)}$ denotes the attention weight derived from Eq.~(\ref{eq:spatial_attention}), and $V_s$ represents the set of spatial points located within the corresponding bounding box region of the search feature map.
The visualization reveals that, regardless of the airplane's grayscale and orientation changes, the search attention consistently focuses on geometrically salient regions of the template, such as the wings and fuselage, rather than merely on texture features.
This strongly demonstrates the effectiveness of TAM in enhancing the model's ability to learn and preserve target geometric structures under complex appearance variations.

\begin{figure}[!t]
    \centering
    \includegraphics[width=0.95\linewidth]{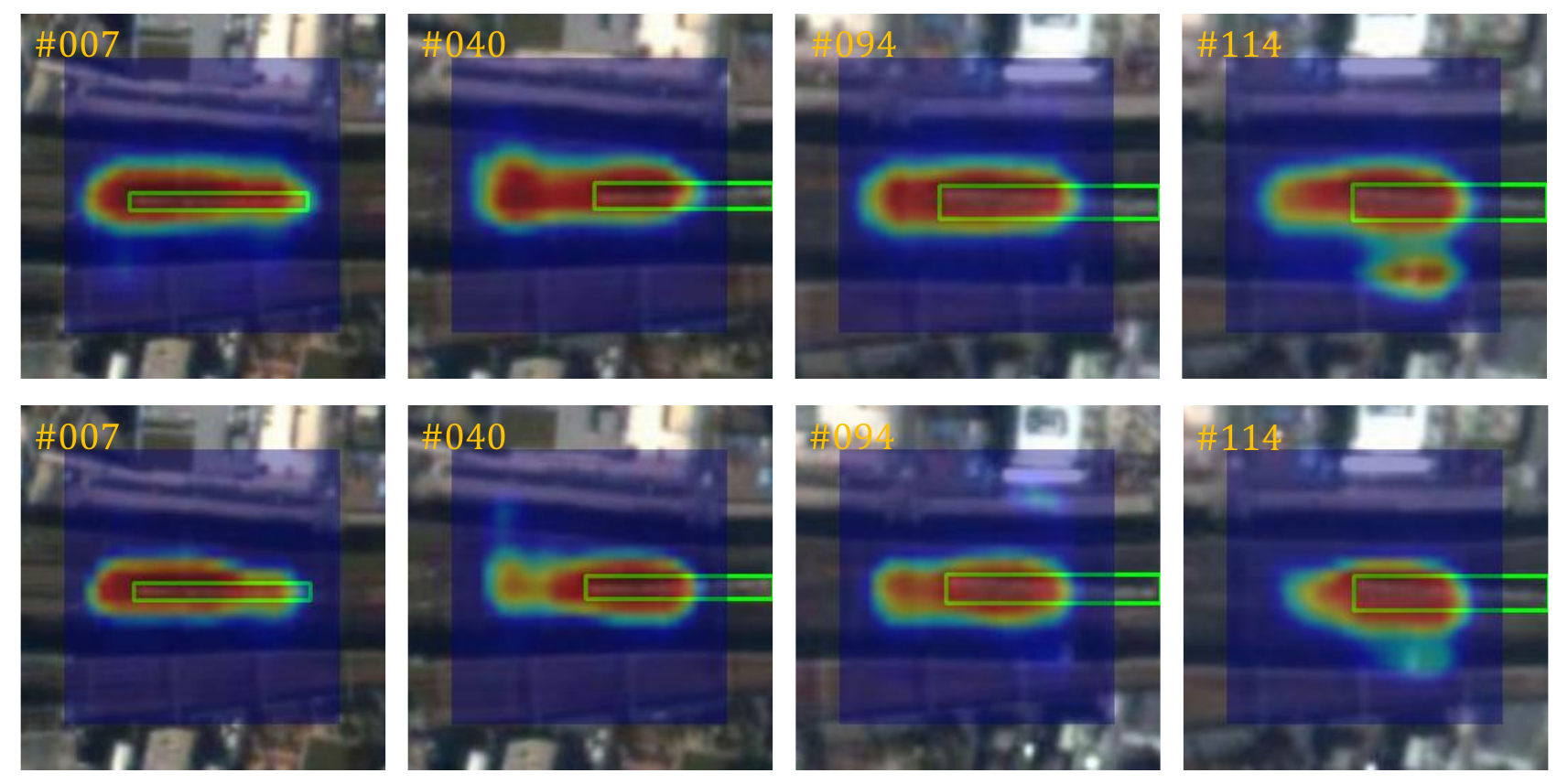}\\[0.5em]
    \includegraphics[width=0.95\linewidth]{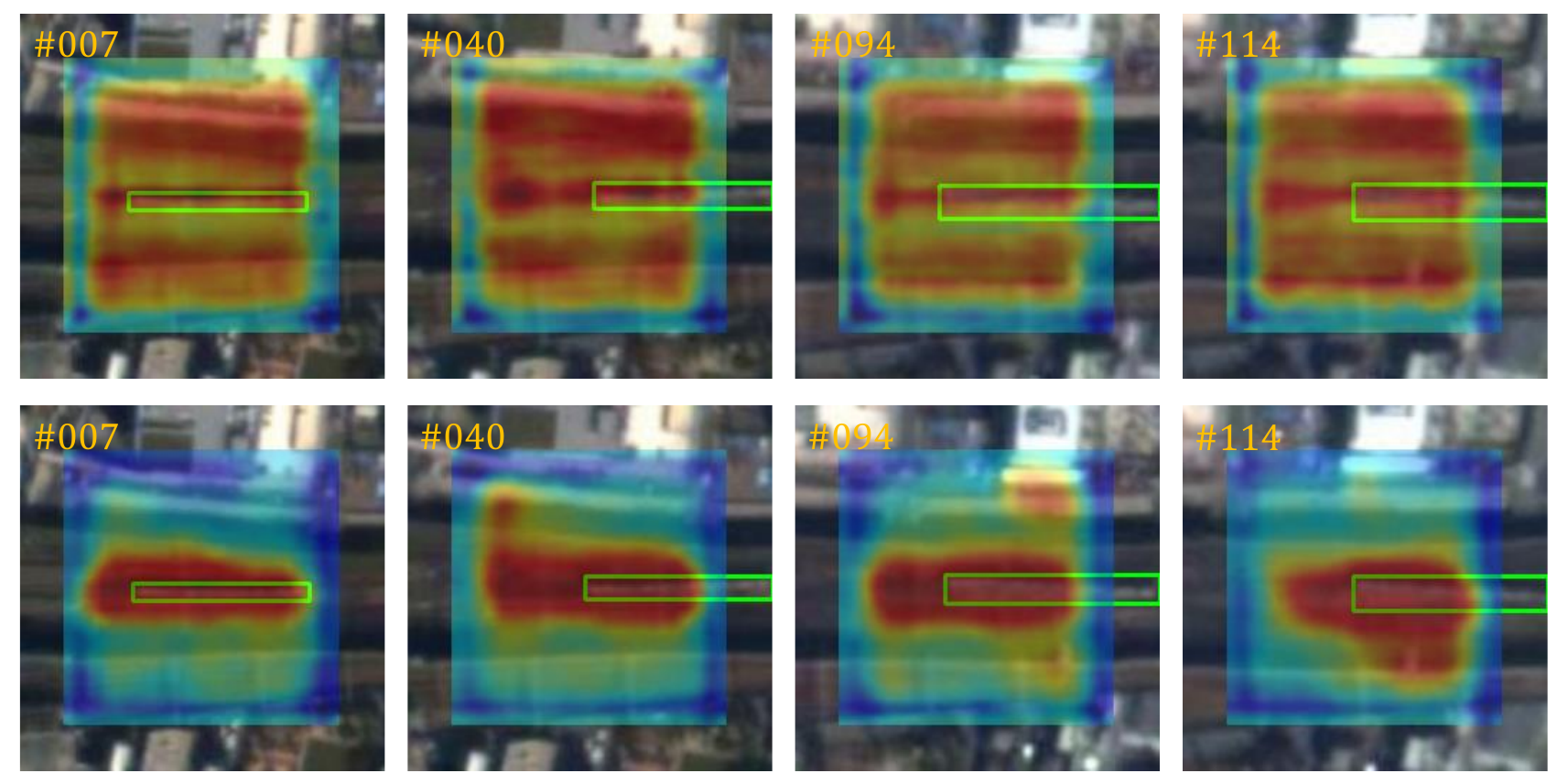}
    \caption{Visualization of the classification and centerness maps during the tracking process on the \texttt{train\_08} sequence. The top two rows show the classification map outputs, while the bottom two rows show the centerness map outputs. For each type of map, the first column presents the results from the baseline model, and the second column shows the outputs from SiamCAR-GCLA. The ground truth is highlighted with a green bounding box.}
    \label{fig:vis_la}
    \vspace{-5pt}
\end{figure}

Moreover, to intuitively demonstrate the effectiveness of our proposed geometry-constrained label assignment strategy, we visualize the classification and centerness outputs during the tracking process.
As shown in Fig.~\ref{fig:vis_la}, we compare the baseline model with our SiamCAR-GCLA.
For targets with large aspect ratios, the baseline model often produces scattered and inaccurate response maps, leading to potential tracking drift.
In contrast, SiamCAR-GCLA dynamically modulates the centerness distribution based on the target's aspect ratio.
This constraint effectively guides the network to concentrate on the most discriminative central regions of the target, resulting in more compact and accurate classification and centerness outputs.
This visual evidence aligns with our quantitative results, confirming that the proposed label assignment strategy significantly enhances tracking robustness for targets with extreme aspect ratios.

\begin{figure*}[!t]
    \centering
    \includegraphics[width=0.95\linewidth]{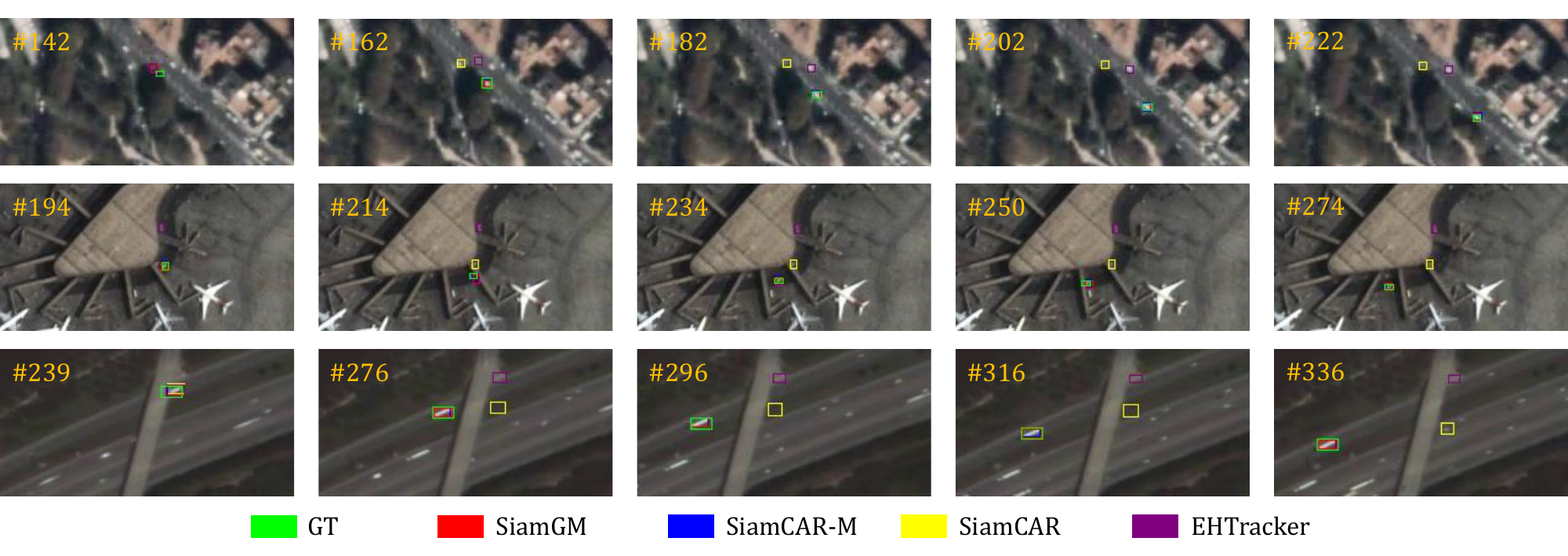}
    \caption{Visualization of the tracking results for three sequences (\texttt{car\_35}, \texttt{car\_58}, and \texttt{car\_61}) under occlusion and tiny object scenarios. The first two rows correspond to POC+TO and FOC+TO attributes, while the third row corresponds to FOC. The red and blue bounding boxes represent the predictions with OMMR, while the green box indicates the ground truth.}
    \label{fig:vis_ommr}
    \vspace{-5pt}
\end{figure*}

Additionally, to further illustrate the practical impact of the OMMR strategy on robust trajectory prediction, we visualize the tracking results of three challenging sequences (\texttt{car\_35}, \texttt{car\_58}, and \texttt{car\_61}), which correspond to the POC+TO, FOC+TO, and FOC attributes, respectively.
As depicted in Fig.~\ref{fig:vis_ommr}, the qualitative frames are sampled at intervals of approximately 20 frames (roughly equivalent to 2 seconds).
In the first two sequences, the extreme tiny scale of the cars, combined with sporadic blockages by trees and building shadows, severely corrupts the visual response maps and induces substantial tracking deviations in the baseline prediction.
However, models with OMMR (indicated by the red and blue bounding boxes) effectively rectify these predictions by dynamically relying on motion vectors, thereby ensuring smooth and continuous tracking.
The third sequence presents a classic full occlusion scenario where the target becomes completely invisible between frames 240 and 275.
While purely appearance-based outputs fail completely during this period, OMMR successfully reconstructs the vehicle's unobservable motion trajectory via long-term linear fitting.
This emphatically demonstrates its powerful capability to bridge severe informational gaps in extreme tracking environments.

%% file: conclusion.tex
\section{Conclusion}
\label{sec:conclusion}

In this paper, we move beyond conventional appearance-centric paradigms to propose SiamGM, redefining SVOT through a unified spatial-temporal perspective.
The core insight of our work is that for blurred and rotated targets, topological structure perception combined with macroscopic shape prior offers a far more resilient representation than traditional raw pixel matching.
Symmetrically, in the temporal inference, dynamically integrating historical motion information serves as a critical strategy, effectively preventing irreversible tracking drift during severe occlusions where visual features fail completely.
Empowered by these designs, our SiamGM achieves state-of-the-art performance on the SatSOT and SV248S benchmarks while operating at a speed far in excess of that required for real-time tracking.

Despite these advancements, certain limitations remain.
First, the improvements yielded by our geometric-topological perception design remain relatively limited for extremely blurred targets, as it is profoundly challenging to fully capture their true morphological structures under severe visual degradation.
Additionally, although our geometry-constrained label assignment significantly mitigates background noise, traditional horizontal bounding boxes still intrinsically restrict the upper limit of localization precision for rotating targets.
Second, the trajectory fitting within our OMMR strategy fundamentally relies on relatively smooth and linear kinematic assumptions.
It may still struggle to accurately predict locations under background jitter or background clutter, and if a target undergoes highly unpredictable motions, our OMMR may fail to provide accurate predictions.

In future work, we plan to integrate object detection and tracking tasks into a unified continuous monitoring framework.
Furthermore, we will explore fine-tuning self-supervised pre-trained foundational models to elevate feature representation capabilities, thereby fundamentally enhancing the robustness of target features against extreme environmental interferences.